\documentclass[letterpaper]{article}
\usepackage{uai2020}
\usepackage[margin=1in]{geometry}

\usepackage{microtype}
\usepackage{graphicx}
\usepackage{booktabs} 
\usepackage{mathtools}
\usepackage{adjustbox}
\usepackage{multirow}
\usepackage{graphicx}
\usepackage{subcaption}
\usepackage{epstopdf}
\usepackage{booktabs} 
\usepackage{amsmath}
\usepackage{amsthm}
\usepackage{amssymb}
\newtheorem{remark}{Remark}
\newtheorem{definition}{Definition}
\newtheorem{fact}{Fact}
\newtheorem{assumption}{Assumption}
\newtheorem{corollary}{Corollary}

\newtheorem{proposition}{Proposition}
\newtheorem{lemma}{Lemma}
\usepackage{bm}
\usepackage{algorithmic}
\usepackage{algorithm}
\usepackage{soul}
\usepackage[xcolor=dvipsnames]{xcolor}
\usepackage{bookmark}
\usepackage{tikz}
\usepackage{pgfplots}
\usepackage{multirow}
\usetikzlibrary{shapes.geometric,arrows,shapes,calc,bayesnet,fit}
\usepackage{longtable}
\usepackage{breqn}
\usepackage{siunitx}
\usepackage{wrapfig}
\usepackage{todonotes}
\usepackage[round]{natbib}

\usepackage[toc,title,page]{appendix}
\usepackage{hyperref}
\usepackage{url}

\hypersetup{
	colorlinks = true,
    urlcolor  = blue,
	citecolor = {violet},
	linkcolor = {violet}}

\DeclareMathOperator*{\argmin}{arg\,min}

\pgfplotsset{compat=1.14}

\usepackage{hyperref}

\usepackage{times}

\title{On Counterfactual Explanations under Predictive Multiplicity}

\author{ {\bf Martin Pawelczyk} \\
University of Tuebingen\\
Tuebingen, Germany 
\And
{\bf Klaus Broelemann}  \\
Schufa AG      \\
Wiesbaden, Germany 
\And
{\bf Gjergji Kasneci}   \\
University of Tuebingen\\
Tuebingen, Germany 
}

\begin{document}

\maketitle

\begin{abstract}
Counterfactual explanations are usually obtained by identifying the \emph{smallest change} made to an input to change a prediction made by a fixed model (hereafter called \emph{sparse methods}). Recent work, however, has revitalized an old insight: there often does not exist one superior solution to a prediction problem with respect to commonly used measures of interest (e.g. error rate). In fact, often multiple different classifiers give almost equal solutions. This phenomenon is known as \emph{predictive multiplicity} \citep{breiman2001statistical,marx2019predictive}. In this work, we derive a general upper bound for the costs of counterfactual explanations under predictive multiplicity. Most notably, it depends on a \emph{discrepancy} notion between two classifiers, which describes how differently they treat negatively predicted individuals. We then compare \emph{sparse} and \emph{data support} approaches empirically on real-world data. The results show that \emph{data support} methods are more robust to multiplicity of different models. At the same time, we show that those methods have provably higher cost of generating counterfactual explanations under one fixed model. In summary, our theoretical and empirical results challenge the commonly held view that counterfactual recommendations should be \emph{sparse} in general.
\end{abstract}

\section{INTRODUCTION}
Counterfactual explanations are usually obtained by identifying the \emph{smallest change} made to an input vector to qualitatively influence a prediction of a pretrained classifier in a positive way; for example, from 'loan rejected' to 'awarded' or from 'high risk of cardiovascular disease' to 'low risk'. But what is a good counterfactual?

\paragraph{A tale of 2 camps.}
The literature commonly agrees that counterfactual explanations mainly serve two purposes \citep{ustun2019actionable,karimi2019model,wachter2017counterfactual}: First, they should help understand a model's \emph{local decision boundary}, answering questions like \emph{"Why does the model give a certain prediction for a given individual?"} (\emph{purpose I}). Second, counterfactual explanations should provide \emph{a recommendation/recommendations} for the individual in question (\emph{purpose II}). Hence, they should give answers to the question \emph{"What is the \emph{smallest change} in inputs an individual needs to make in the future to receive a desired outcome?"}. In this work, we focus on purpose II and analyze counterfactual explanations from the recommendation perspective.

Often there exist several ways to make a reasonable recommendation. So, what constitutes a reasonable recommendation? We roughly split the current literature into two camps. The first line of work (we call them \emph{sparse} counterfactuals) assumes that counterfactual recommendations with minimal change in $\ell_p$-norm are most desirable \citep{wachter2017counterfactual,grath2018interpretable,russell2019efficient,ustun2019actionable,laugel2017inverse,karimi2019model,tolomei2017interpretable}. 

The second camp (henceforth called \emph{(Data) Support} counterfactuals) suggests that the norm should receive second order importance when generating counterfactual explanations. Instead, it would be more desirable to generate counterfactual recommendations that are close to correctly classified observations from the desired class and semantically meaningful \citep{laugel2019dangers,laugel2019issues,pawelczyk2019learning,joshi2019towards,mahajan2019preserving}. We give exact definitions of both types in section \ref{sec:related_work}. Finally, a recent line of work considers causal interventions to generate counterfactual explanations \citep{karimi2020algorithmic}.

All aforementioned works assume that the pretrained classifier is given and that there exist no uncertainty as to whether it is the best possible classifier or whether it will remain the classifier of choice over time. Counterfactual recommendations are then usually generated with reference to this 'best' pretrained model. 

\begin{table}[htb!]
    \centering
    \begin{adjustbox}{width=\columnwidth,center}
    \begin{tabular}{ccccc}
         \toprule
          Proposal & Input subset & current value & & required   \\
         \cmidrule(lr){1-1} \cmidrule(lr){2-2} \cmidrule(lr){3-5}
          1 & \# \texttt{credit cards} & 5 & $\xrightarrow{}$ & 3  \\
         \cmidrule(lr){1-5}
         2 & \texttt{current debt} & \$3250 & $\xrightarrow{}$ & \$1000 \\
         \cmidrule(lr){1-5}
         \multirow{2}{*}{3} & \texttt{has savings account} & 0 & $\xrightarrow{}$ & 1 \\
         & \texttt{has retirement account} & 0 & $\xrightarrow{}$ & 1 \\
         \bottomrule
    \end{tabular}
    \end{adjustbox}
    \caption{Stylised example from an individual who was denied credit by a fixed classifier $f$, i.e. $\text{sign}(f(\bm{x}_{current})) = -1$, and three different associated counterfactual recommendations, i.e. $\text{sign}(f(\bm{x}_{required})) = +1$. \textbf{The difference between the current values and the required values are the costs of counterfactual recommendations}. Example taken from \citet{ustun2019actionable}.}
    \label{tab:my_label}
\end{table}

\paragraph{\textbf{Counterfactuals under model multiplicity.}}
Recent work has revitalized an old insight \citep{marx2019predictive}: there often does not exist one superior solution to a prediction problem with respect to commonly used measures of interest (e.g. error rate). In fact, often multiple different models give almost equal solutions. This phenomenon is known as \emph{predictive multiplicity} \citep{marx2019predictive, breiman2001statistical,mccullagh1989generalized} and in this work we argue that it should shape our understanding of how counterfactual recommendations are generated.

Admitting the existence of several well performing models for the very same prediction task calls the entire business of generating counterfactual recommendations for one particular model into question. Or, as Leo Breiman already put it \citep{breiman2001statistical}: \emph{"[...] if there exist several equally good models for a given dataset [sic], each of which provides a different explanation of the data-generating process, then how can we tell which one is correct?"}. 

\citet{barocas2019hidden} identify 4 hidden assumptions underlying the generation of counterfactual explanations: (\textbf{A1}) \emph{The underlying model is stable over time}. (\textbf{A2}) \emph{Explanations can be offered without regard to decision making in other areas of people's lives}. (\textbf{A3}) \emph{Counterfactual explanations map to real world actions}. (\textbf{A4}) \emph{Inputs can be made commensurate by looking at the training data}. In this work, we will investigate the effect of assumption \textbf{A1} for counterfactual explanations in theory and in practice.

To get a better understanding for the underlying problem, let us consider the two following scenarios:
\begin{itemize}
    \item[(a)] The decision maker decides to change the deployed model $f$ at some point $\tau$. However, up to $\tau$, $f$ was used to generate counterfactual recommendations. Will the recommendations still lead to the desired outcome under the competing model $g$? What are the expected additional costs due to the introduction of $g$? Will the cost depend on whether we use a \emph{sparse} or \emph{Data Support} counterfactual explanation machine?
    \item[(b)]  The decision maker is unsure about the correct classifier $f$ and multiple models give almost identical hold-out test error. Will the explanations $E(\bm{x};f)$ based on the model $f$ generalize to a family of competing models $g$?
\end{itemize}
Point (b) refers to a concept usually known as 'researcher/practitioner degrees of freedom'. It describes that it is often not very clear why a certain classifier was chosen from a set of (potentially equally well performing) classifiers. On an individual end-user level, those choices make a difference and can have vast consequences. For example, they can determine whether someone gets a loan or not or whether a decision should be revised or not.

While there has been a sharp recent increase in the availability of methods that attempt to generate counterfactual recommendations (see section \ref{sec:related_work}), there exists remarkably little work regarding their cost guarantees. At the same time, such guarantees might be a crucial element when deciding which method (\emph{sparse} vs. \emph{data support}) should be deployed in practice for consequential decisions with humans in the loop. This work attempts to close this gap.

\paragraph{\textbf{Our contributions}.} In this paper, we challenge commonly held assumptions in the field of counterfactual explanations. We summarize our key contributions briefly:
\begin{itemize}
    \item \textbf{Relating the costs of \emph{Sparse} and \emph{Data Support} counterfactuals.}
    We theoretically relate the cost of \emph{sparse} and \emph{Data Support} counterfactual recommendations. When the classifier is fixed, our result shows that \emph{sparse} recommendations are provably less costly than those with \emph{Data Support}. Under model multiplicity, we derive conditions which depend on the relative costs of both methods.
    \item \textbf{Cost guarantees for counterfactuals under model multiplicity}. We derive an upper bound on the cost of counterfactual explanations under model multiplicity. Our upper bound is stated in terms of the risk of both classifiers and most notably depends on how differently both classifiers assign negative predictions. Our result challenges the commonly held view that a counterfactual recommendations should have the \emph{lowest possible cost} in general.
    \item \textbf{Empirical evaluation of the result.} Empirically, we compare \emph{Data Support} and \emph{Sparse} methods. Given one fixed classifier $f$, the \emph{Data Support} recommendations are theoretically and empirically more costly, however, they are empirically more invariant to multiplicity of different models than \emph{sparse} recommendations and result in semantically sensible recommendations.
\end{itemize}

\paragraph{\textbf{Structure}.} In section \ref{sec:related_work} we give a categorization of different approaches. Section \ref{sec:theoretical_results} contains theoretical cost guarantees, where we briefly discuss their implications. In section \ref{sec:evaluation}, we describe the compared models and evaluate them both quantitatively with respect to \emph{cost} and \emph{invariance to predictive multiplicity}, and qualitatively with respect to their semantics. Finally, section \ref{sec:discussion} concludes.


\section{RELATED WORK}\label{sec:related_work}
We denote the $d$-dimensional feature space as $\mathcal{X}=\mathbb{R}^d$ and the feature vector for observation $i$ by $\bm{x} \in \mathcal{X}$ and the $j$-th dimesn denotes the  The labels corresponding to the $i$-th observation are denoted by $y \in \mathcal{Y} = \{-1,+1\}$. Moreover, we assume two given pretrained, not identical, classifiers $f,g: \mathbb{R}^d \xrightarrow{} \mathbb{R}$. Depending on the sign of $f(\bm{x})$ or the sign of $g(\bm{x})$ instances are classified. The goal is to find a counterfactual recommendation system for a given $f$, $E_f:\mathcal{X} \xrightarrow{ }\mathcal{X}$, generating counterfactuals $E(\bm{x};f) = \tilde{\bm{x}}$, such that $\text{sign}(f(\bm{x})) \neq \text{sign}(f(E(\bm{x};f)))$. We also introduce the following sets: 
\begin{align*}
    H^{+}_f = \{\bm{x} \in \mathcal{X}: f(\bm{x}) > 0 \},  H^{-}_f = \{\bm{x} \in \mathcal{X}: f(\bm{x}) \leq 0 \} \\
    H^{+}_g = \{\bm{x} \in \mathcal{X}: g(\bm{x}) > 0 \},  H^{-}_g = \{\bm{x} \in \mathcal{X}: g(\bm{x}) \leq 0 \} \\
    D^{+} = \{\bm{x} \in \mathcal{X}: y = +1 \},   D^{-} = \{\bm{x} \in \mathcal{X}: y = -1 \}
\end{align*}

\paragraph{\textbf{Exploring the local decision boundary}.}
This line of work targets \emph{purpose I} only. These approaches are based on perturbations and attempt to explain the sensitivity of a machine learning model to changes in its inputs by modelling the impact of local perturbations 
\citep{ribeiro2018anchors,adler2018auditing,fong2017interpretable}. Examples of perturbation-based approaches are LIME \citep{ribeiro2016should} and SHAP \citep{lundberg2017unified}. However, none of them attempts to make recommendations for users who are directly affected by the classifications.

\begin{figure*}[htb!]
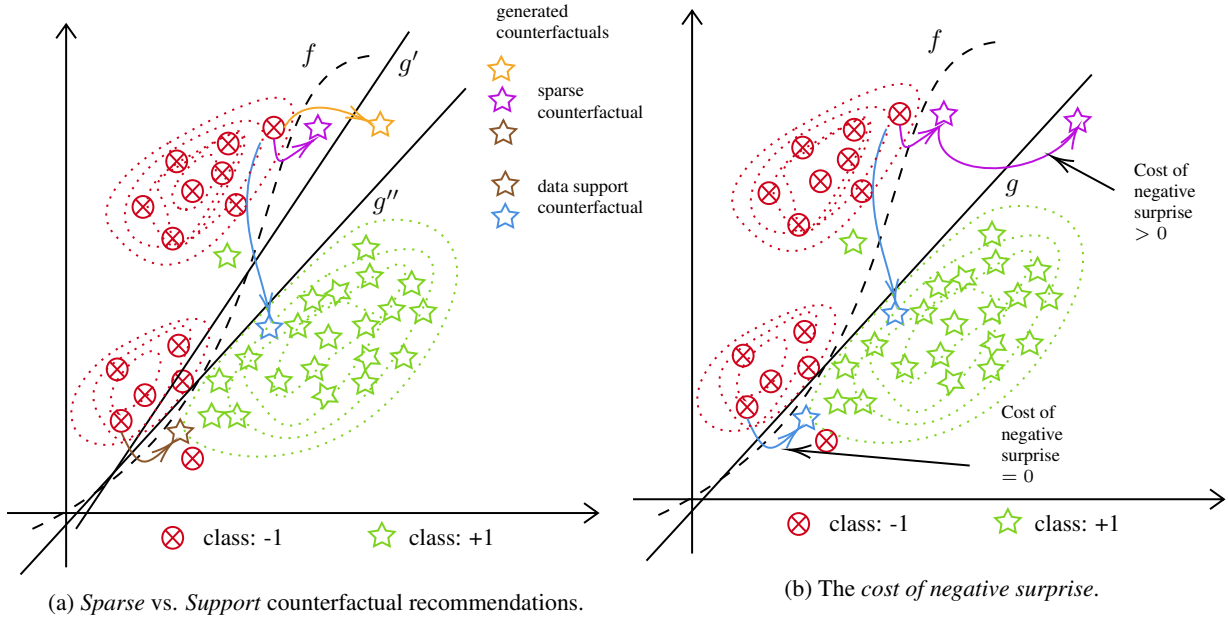

\tikzset{every picture/.style={line width=0.75pt}} 
\begin{subfigure}[c]{0.5\textwidth}
\tikzset{every picture/.style={line width=0.75pt}} 

\subcaption{\emph{Sparse} vs. \emph{Support} counterfactual recommendations.}
\end{subfigure}
\begin{subfigure}[c]{0.5\textwidth}

\tikzset{every picture/.style={line width=0.75pt}} 

%
\subcaption{The \emph{cost of negative surprise}.}
\end{subfigure}
\caption{(\textbf{a}) \emph{Data support counterfactual} (def.\ \ref{def:rec_data_support}) recommendations are more costly than \emph{sparse counterfactual} recommendations (def.\ \ref{def:rec_unrestricted}). See also proposition \ref{proposition:oracle_cost}. (\textbf{b}) Suppose we made counterfactual recommendations under model $f$. If at some point $\tau$ we changed from $f$ to $g$, then the \emph{cost of negative surprise} is 0 for data support counterfactual recommendations while it is positive for sparse counterfactuals. Sparse counterfactual recommendations are more vulnerable to classifier uncertainty or classifier changes over time. Although more costly in the first place, \emph{data support} counterfactuals are more transferable across different classifiers, i.e. they tend to have \emph{lower cost of negative surprise.}}
\label{fig:visualizing_sparse_negative}
\end{figure*}

\paragraph{\textbf{Counterfactual explanations}.} These works target both \emph{purposes I and II}. Approaches dealing with tabular data rely on solving integer programming optimization problems \citep{ustun2019actionable,russell2019efficient}, use decision tree based classifiers \citep{tolomei2017interpretable}, satisfiability modulo theory \citep{karimi2019model} or use data density approximation (via variational autoencoders) \citep{joshi2019towards,pawelczyk2019learning}. Other approaches ignore tabular data entirely \citep{grath2018interpretable,laugel2017inverse}, but at least allow for conditionally immutable features (e.g. has a PhD) \citep{lash2017generalized}. To produce counterfactuals that take on reasonable values (e.\ g. non negative values for wage income) most approaches let decision makers specify the set of features and their respective support subject to change. We next aim to categorize most of the aforementioned approaches.

\subsection{Sparse Approaches}
\begin{definition}{\textbf{Sparse counterfactual recommendation.}}
Given inputs $\bm{x} \sim p_{data}$, a binary classifier $f(\bm{x})$ and a set of all possible counterfactual explanations, $\mathcal{E}_{S} = \{\tilde{\bm{x}}:\ \text{sign}(f(\tilde{\bm{x}})) = +1\}$, a sparse counterfactual recommendation is defined as $\bm{c}_{S}=\argmin_{\tilde{\bm{x}} \in \mathcal{E}_{S}}$ $\lVert \tilde{\bm{x}} - \bm{x} \rVert_p$.
\label{def:rec_unrestricted}
\end{definition}

Owing to interpretabiliy, $p$ is usually $1$ or $2$. Several works have been put forth, relying on a variant of this definition \citep{laugel2017inverse, karimi2019model, grath2018interpretable, russell2019efficient, ustun2019actionable, wachter2017counterfactual,lash2017generalized,10.1145/3351095.3372850}. In fact, a subset of these works additionally considered to restrict $\mathcal{E}_{S}$ further; for example some suggest to favour explanations over inputs that have shown to vary much in the past \citep{wachter2017counterfactual}, aim at generating diverse recommendations \cite{10.1145/3351095.3372850} or allowed for having immutable inputs that could not be changed (e.g.\ \emph{Gender}, \emph{Age}) while searching for possible counterfactual recommendations \citep{lash2017generalized,ustun2019actionable}.

\subsection{Data Support Approaches}
\begin{definition}{\textbf{(Data) Support counterfactual recommendation}.}
Given inputs $\bm{x} \sim p_{data}$, a binary classifier $f(\bm{x})$ and a set of all admissible counterfactual explanations, $\mathcal{E}_{D} = \{\tilde{\bm{x}}:\ \text{sign}(f(\tilde{\bm{x}}) = +1\  \text{ s.t. } ~ p_{data}(\tilde{\bm{x}}) > 0 \}$, a data supported counterfactual recommendation is defined as $\bm{c}_{D}= \argmin_{\tilde{\bm{x}} \in \mathcal{E}_{D}} \lVert \tilde{\bm{x}} - \bm{x} \rVert_p$.
\label{def:rec_data_support}
\end{definition}
Definition \ref{def:rec_data_support} essentially demands that counterfactual recommendations should be supported by the true data distribution $p_{data}$. This is what is meant by \emph{data support} and it comes at a cost since it is easy to see that $\bm{c}_{D}\geq \bm{c}_{S}$. In proposition \ref{proposition:oracle_cost} below we refine this statement. Additionally, consider figure \ref{fig:visualizing_sparse_negative} for an example. Of course, in practice we do not know $p_{data}$ and therefore an explainability generator $E(\bm{x};f)$ would need to take density estimation into account. Notice that this notion is also distinct from \emph{actionability} \citep{ustun2019actionable} or \emph{plausibility} \citep{karimi2019model}. They only demand that immutable inputs shall not be changed and that columns of $\tilde{\bm{x}}$ lie individually in a reasonable range. Per se, this does not imply $p_{data}(\tilde{\bm{x}})>0$. \citet{laugel2019issues} suggested density based evaluation measures to approximate whether $p_{data}(\tilde{\bm{x}})>0$ holds. A small collection of works has devised methods to generate \emph{data support counterfactual recommendations} \citep{joshi2019towards,pawelczyk2019learning,mahajan2019preserving} using variational autoencoders \citep{kingma2013auto,nazabal2018handling}. 

In light of the fact that counterfactual recommendation machines could have a huge impact on individuals' lives, there there exists remarkably little work regarding cost guarantees. With respect to our theoretical results the most relevant work is by \citet{ustun2019actionable}. Proposition \ref{proposition:expected_cost_multiplicity} is more general (see remark \ref{remark:one}) since it considers the case of predictive multiplicity and nonlinear classifiers. In fact, it includes their result as a special case when the considered classifiers $f$ and $g$ coincide and are both linear. To the best of our knowledge, our work is the first that aims at relating the cost of \emph{sparse} and \emph{data support} counterfactuals.


\section{COST GUARANTEES}\label{sec:theoretical_results}
 \subsection{Relation between \emph{Sparse} and \emph{Data support} Costs}

\citet{pawelczyk2019learning} establish empirically that there exists a trade-off between low-cost recommendations (\emph{sparse}) and those with data support (they call them \emph{attainable}). Here we give a theoretical underpinning of this empirical observation.

Suppose $\bm{x}$ is generated by a generative model $h$ such that $h(\bm{z})=\bm{x}$, where $\bm{z} \in \mathcal{Z}=\mathbb{R}^k$ are latent codes with $k<d$. As an example, $\bm{z}$ could be standard normal distributed. If the generative model was an autoencoder, this amounts to having a perfect encoder and decoder. As in \citet{pawelczyk2019learning}, we consider the following explanation mechanism to devise counterfactual recommendations via a nearest neighbour search in latent space: 
\begin{equation}
   f(h(\tilde{\bm{z}})) \text{ where } \tilde{\bm{z}} = \argmin_{\bm{z}} \lVert h(z) - x \rVert.
   \label{eq:latent_nn}
\end{equation}
Then the following result holds.
\begin{proposition}[Oracle cost inequality]\label{proposition:oracle_cost}
The cost relation between \emph{sparse} counterfactual recommendations and \emph{data supported} counterfactual recommendations adheres:
\begin{equation*}
    \bm{c}_{D}(\tilde{\bm{z}}) \leq 2 \cdot \bm{c}_{S},
\end{equation*} 
\label{proposition:oracle_inequality}
\end{proposition}
where \emph{S} and \emph{D} abbreviate \emph{sparse} and \emph{data support}, respectively. The proof can be found in appendix \ref{appendix:oracle_cost}. If we wish to obtain counterfactual recommendations with data support,
proposition \ref{proposition:oracle_cost} suggests that there exists an extra cost, relative to the \emph{sparse} counterfactual recommendations. In practice, however, a generative model is used for which the encoder and decoder parameters have to be estimated adequately. Therefore, the cost difference is likely to be higher since neither the encoder nor the decoder work perfectly. Our experiments in section \ref{sec:evaluation} consolidate our findings.

\subsection{Cost of Counterfactual Multiplicity}
We start by stating the general objective. The goal is to find a minimal cost action $c^*$ which alters the given classifiers' predicted labels from $\text{sign}((f(x)) = -1$ and $\text{sign}(g(x))=-1$ to $+1$. More formally, we seek:
\begin{equation}
    \begin{split}
& c^*(f,g) = \argmin_{c \in \mathbb{R}^d} \lVert c \rVert \text{ s.t.\ } \\
& \text{sign}(f(x+c)) = +1 ~ \wedge ~ \text{sign}(g(x+c)) =+1.
    \label{eq:cost_multiplicity}
\end{split}
\end{equation}
Next, we state the main assumption used in proposition \ref{proposition:oracle_inequality}.
\begin{assumption}[\citet{pang1997error}] \label{ass:distance}
There exist $\alpha >0$ and $0\leq \gamma \leq 1$ such that, for all $\bm{x}$,
\begin{align*}
    dist(x, H_f^{+} \cap H_g^{+}) &  \leq \alpha \cdot \text{max}\{0,max(-f(x), -g(x))\}^{\gamma}, \\
    dist(x, H_f^{+} \cap H_g^{-}) & \leq \alpha \cdot \text{max}\{0,max(-f(x), +g(x))\}^{\gamma}, \\
    dist(x, H_f^{-} \cap H_g^{+}) & \leq \alpha \cdot \text{max}\{0,max(+f(x), -g(x))\}^{\gamma}, \\
    dist(x, H_f^{-} \cap H_g^{-}) & \leq \alpha \cdot \text{max}\{0,max(+f(x), +g(x))\}^{\gamma},
\end{align*}
where $dist(x,H) = \underset{s \in H}{\text{min}} \{ \lVert \bm{x}-\bm{s} \rVert\}$.
\end{assumption}
The assumption states that a given point $\bm{x}$ is bounded by the so-called residual. We assume the residual provides a reasonable way to bound the distance from $\bm{x}$ to a point $\bm{s} \in H$ classified as $y = 1$ or $y=-1$. We proceed to define the quantity for which we give an upper bound.
\begin{definition}[Cost of counterfactual multiplicity]\label{def:cost_multiplicity}
The expected cost of counterfactual explanations under classifier multiplicity for classifiers $f: \mathbb{R}^d \xrightarrow{}  \mathbb{R}$ and $g: \mathbb{R}^d \xrightarrow{} \mathbb{R}$ is defined as,
\begin{align*}
    \overline{cost}(f,g)_{H_{f}^{-} \cup H_{g}^{-}} & = \mathbb{E}_{H_{f}^{-} \cup H_{g}^{-}}[c^*(f,g)],
    \label{eq:def_counterfactual_multiplicity}
\end{align*}
\end{definition}
where the expectation is taken over the distribution of $\bm{x} \in H_{f}^{-} \cup H_{g}^{-}$. Analogs can be defined in which costs can be computed with respect to classifiers $f$ or $g$ only. For example, for the classifier $f$ we would obtain $\overline{cost}(f)_{H_{f}^{-}} = \mathbb{E}_{H_{f}^{-}}[c^*(f)]$, where the objective in \ref{eq:cost_multiplicity} would need to be altered appropriately. Definition \ref{def:cost_multiplicity} asks to find the expected minimum cost of counterfactual recommendations when we have to satisfy the constraint set out by two classifiers (see \eqref{eq:cost_multiplicity}).

\begin{proposition}[Bounding costs of counterfactual multiplicity]\label{proposition:expected_cost_multiplicity}
Given assumption \ref{ass:distance}, the cost of counterfactual multiplicity under classifiers $f$ and $g$, $\overline{cost}(f,g)_{H_{f}^{-} \cup H_{g}^{-}}$, is bounded from above such that,
\begin{align}
    \begin{split} 
            & \overline{cost}(f,g)_{H_{f}^{-} \cup H_{g}^{-}} \leq \alpha \cdot 8^{1-\gamma}  \\
      \cdot \Big[ & \underbrace{2 \cdot R_{H_f^-}(f)\cdot  c^{max}_{H_f^-}(f) + 2 \cdot R_{H_g^-}(g) \cdot c^{max}_{H_g^-}(g)}_{\text{maximum risk of f and g}} \\
    & + \underbrace{\pi_f \cdot c_{D^+}(f) + \pi_g \cdot  c_{D^+}(g)}_{\text{false negative rates of f and g}} \\
    & - \underbrace{(1-\pi_f) \cdot  c_{D^-}(f)  - (1-\pi_g)  \cdot  c_{D^-}(g)}_{\text{true negative rates of f and g}} \\
    & + \underbrace{\mathbb{E}_{H_f^- \cup H_g^-}[|f(x)-g(x)|]}_{\substack{\text{$\bm{\Delta}(f,g)$: Discrepancy of f and g} \\ \text{over neg. classified individuals}}}\Big]^\gamma, \text{where}
    \end{split}
\end{align}
\begin{itemize}
    \item $c_{D^+}(f) = \mathbb{E}_{H^-_f\cap D^+} [f(x)]$ is the expected cost of counterfactual recommendations for individuals with $\bm{x} \in H^-_f \cap D^+$ (\emph{false negative predictions});
    \item $c_{D^-}(f) = \mathbb{E}_{H^-_f\cap D^-} [f(x)]$ is the expected cost of counterfactual recommendation for individuals with $\bm{x} \in H^-_f \cap D^-$ (\emph{true negative predictions});
    \item $c^{max}_{H_f^-}(f)= \max_{x \in H^-_f} |f(x)|$ denotes the max.\ cost of counterfactual recommendation for classifier $f$;
    \item $\pi_f = Pr_{H_f^-}(y=1)$ is the false-omission rate of classifier $f$;
    \item $R_{H_f^-} (f) = \pi_f Pr_{H^-\cap D^+}(f(x)\leq 0) + (1-\pi_f) Pr_{H^-_f\cap D^-}(f(x)>0)$ stands for the risk of classifier $f$ for $x\in H^-_{f}$.
\end{itemize}
\end{proposition}
Analogs can be defined for classifier $g$ and the proof is given in appendix \ref{appendix:expected_cost_multiplicity}. The expected cost of counterfactual explanations under predictive multiplicity does not directly depend on the overall classification error rate, but instead \emph{focuses on those individuals for whom we made negative predictions}. 

\begin{figure}[htb!]
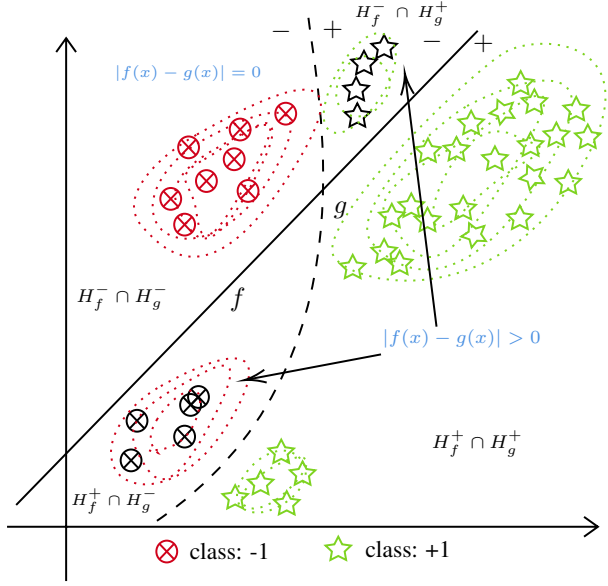

    \centering
\tikzset{every picture/.style={line width=0.75pt}} 


    \caption{\textcolor{blue}{Discrepancy of $f$ and $g$}. For the area $H_f^- \cap H_g^-$ (upper left corner) the discrepancy $\lvert f(x)-g(x)\rvert$ is 0: both models agree that they should be classified negatively. The \textcolor{red}{red junctions} in the upper left corner can neither be moved to $H_f^+ \cap H_g^-$ nor to $H_f^- \cap H_g^+$. Since these areas are not admissible we have to pay an extra price for them to be moved to $H_f^+ \cap H_g^+$. This intuition is captured by the $\Delta(f,g)$ term in proposition \ref{proposition:expected_cost_multiplicity}.} 
    \label{fig:discrepancy}
\end{figure}

We would like to highlight the discrepancy term in proposition \ref{proposition:expected_cost_multiplicity}. Consider figure \ref{fig:discrepancy} for a more illustrative explanation of this term. Although the areas $H_f^+ \cap H_g^-$ and $H_f^- \cap H_g^+$ are already covered by the false and true negative rates of both classifiers, the discrepancy term counts them again. Intuitively, this is due to the fact that the red junctions in the upper left corner can neither be moved to $H_f^+ \cap H_g^-$ nor to $H_f^- \cap H_g^+$.

We would now like to take a step back and highlight some noteworthy real-world implications of this result:

\begin{itemize}
\item \textbf{Challenge minimal cost recommendations}. When assumption \textbf{A1} (\citet{barocas2019hidden}; also mentioned in the introduction) is violated, our analysis suggests that finding sparse, minimal cost counterfactual recommendations with respect to a fixed classifier $f$ fails to reflect the real expected cost of counterfactual recommendations.
\item \textbf{Distort trust in automated ML}. If declined end users are initially issued a list of recommended feature changes and those changes turn out to be more costly due to, say, model updates over time, then this can severely distort trust in automated decision making systems. To the best of our knowledge, a legal framework for such cases does not exist, yet.
\end{itemize}

\begin{remark}
If we take $\gamma = 1$ and $f$ and $g$ coincide, i.e. $f(x)=g(x) ~ \forall x$, then our result recovers theorem 3 in \cite{ustun2019actionable}, where $\gamma = 1$ corresponds to the case where were we look at linear classifiers. 
\label{remark:one}
\end{remark}

Next we  evaluate under which conditions any of the existing methods (\emph{sparse} vs.\ \emph{data support}) generate more robust counterfactual recommendations.

\subsection{Relating the Cost of Negative Surprise for \emph{Sparse} and \emph{Data Support} Counterfactuals}
We would like to find out what the additional cost induced by the classifier $g$ would be. We call it the cost of \emph{negative surprise}. Intuitively, it measures whether individuals subjected to a particular recommendation method (say \emph{sparse} vs.\ \emph{data support} recommendations) should be worried that their recommendation would change under a different classifier. If the classifiers $g$ and $f$ were to coincide for all instances $\bm{x}$, then the \emph{cost of negative surprise} to all individuals would be 0 since no individual would need to exert additional effort/cost to satisfy a new constraint, which is illustrated in the right panel of figure \ref{fig:visualizing_sparse_negative}.

\begin{definition}[Inverse cost of negative surprise]
The normalized inverse cost of negative surprise under model multiplicity for classifiers $f: \mathbb{R}^d \xrightarrow{}  \mathbb{R}$ and $g: \mathbb{R}^d \xrightarrow{} \mathbb{R}$ under method $M = \{D,S\}$ is defined as:
\begin{align*}
    \overline{s}(f,g)_M & = \left[\frac{\mathbb{E}_{H_{f}^{-} \cup H_{g}^{-}}[c^*(f(x),g(x))]_M}{\mathbb{E}_{H_{f}^{-}}[c^*(f(x))]_M}\right]^{-1} \in (0,1].
\end{align*}
\label{def:negative_surprise}
\end{definition}
The inverse cost is a measure of invariance of a counterfactual recommendation to different classifiers and ideally evaluates to 1. This happens when $\mathbb{E}_{H_{f}^{-} \cup H_{g}^{-}}[c^*(f,g)]_M = \mathbb{E}_{H_{f}^{-}}[c^*(f)]_M$, that is, in expectation there will be no additional changes to the cost of counterfactual recommendations due to the introduction of a competing classifier $g$. 

\begin{remark}
Definition \ref{def:negative_surprise} appears cumbersome, however, it allows us to use a lower bound for $\mathbb{E}_{H_{f}^{-} \cup H_{g}^{-}}[c^*(f,g)]$, which depends on $\mathbb{E}_{H_{f}^{-}}[c^*(f)]$ and $\mathbb{E}_{H_{g}^{-}}[c^*(g)]$. 
\end{remark}

\begin{proposition}[Negative surprise for sparse and data support recommendations]
For simplicity of the statement, suppose $\gamma = 1$. If $\mathbb{E}_{H_f^- \cup H_g^-}[|f(x)-g(x)|]_{S} = \mathbb{E}_{H_f^- \cup H_g^-}[|f(x)-g(x)|]_{D}$ and
\begin{equation*}
    \frac{\mathbb{E}_{H_{g}^{-}}[c^*(g)]_{D}}{\mathbb{E}_{H_{f}^{-}}[c^*(f)]_{D}} < \frac{\mathbb{E}_{H_{g}^{-}}[c^*(g)]_{S}}{\mathbb{E}_{H_{f}^{-}}[c^*(f)]_{S}}.
\end{equation*}
then we must have that:
\begin{equation*}
1 \geq \overline{s}(f,g)_{S} > \overline{s}(f,g)_{D}.  
\end{equation*}
\label{proposition:relation_negative_surprise}
\end{proposition}
The proof of Proposition \ref{proposition:relation_negative_surprise} can be found in appendix \ref{appendix:relation_negative_surprise}. It suggests a direct way to check whether counterfactual suggestions generated by the \emph{sparse} methods are less prone to negative surprise than those generated by the \emph{Data Support} camp.
Note that the \emph{sparse} explanation method could trivially satisfy this condition by generating high cost recommendations for the classifier $g$. 
For example, a \emph{sparse} method could push the \textcolor{red}{red junctions} in the upper left of figure \ref{fig:discrepancy} all the way to the bottom right of the plot. However, by definition \ref{def:rec_unrestricted} \emph{sparse} explanations mechanisms are not set out to do so. It would indeed be counterproductive for the generation of counterfactual recommendations that end-users can realistically translate into lived realities.

This result implies an interesting question (for future research): \emph{Can we generate invariant counterfactual recommendations with minimal costs}? In the following, we do not suggest a new way to do so, but we evaluate whether existing methods already do.


\section{EXPERIMENTS}\label{sec:evaluation}
\paragraph{\textbf{Data sets}.}
We conduct extensive quantitative and qualitative evaluations on two different realistic classification settings: (i) 'Give Me Some Credit' and (ii) HELOC.
\paragraph{\textbf{``Give Me Some Credit''}.} This data set contains 10 inputs and 8 are related to the individual's financial history. We assume that the inputs are mutable and their types are \emph{count} and \emph{positive continuous}, respectively. The remaining 2 features, $\emph{age}$ and \emph{\# dependencies}, are immutable. 

\paragraph{\textbf{HELOC}.} This data set has 23 inputs of which all of them are related to the individual's financial history. All the inputs are treated as count variables. We treat the \emph{ExternalRiskEstimate}, \emph{MSinceOldestTradeOpen} and \emph{AverageMInFile} as immutable since they are not under the individual's direct control. The remaining inputs are treated as \emph{mutable}. The data set originally holds 10000 observations, but after dropping observations with missing instances we are left with $n = 8291$.

\paragraph{\textbf{Methods}.} We choose three methods and compare across three dimensions. First, what is the associated cost of the generated counterfactual recommendations. Second, what is the individual cost of negative surprise, i.e. how well do the generated counterfactual explanations generalize to other models? Third, do the generated recommendations make semantically sense?

(\emph{Sparse methods}) The first chosen method is classifier agnostic and conducts a greedy nearest-neighbour search. It chooses the closest counterfactual recommendations measured by the $\ell_2$-norm \citep{laugel2017inverse} (GS). The second method was suggested by \citet{ustun2019actionable} (AR).  They use integer programming tools subject to cost function \eqref{eq:total_shift} and \eqref{eq:max_shift}. While their method is restricted to linear classifiers, it also works for tabular data. (\emph{Data Support methods}) The last method is classifier agnostic and was concurrently suggested by \citet{joshi2019towards,pawelczyk2019learning,mahajan2019preserving}. We use the method as suggested in \citet{pawelczyk2019learning} (OURS).They use a special type of variational autoencoder (VAE) \citep{kingma2013auto,nazabal2018handling} for counterfactual search. The idea is to train a (V)AE that deals well with tabular data and to leverage the latent space representation to search for counterfactual recommendations.

\subsection{The (Local) Cost of Negative Surprise}
In this section, we investigate different models' ability to generate counterfactual recommendations that generalize well across different classifiers. If they generalize well, then they have a low cost of negative surprise. To do so, we distinguish the following two cases: (a) Holding the hypothesis class fixed, will the initially generated counterfactual recommendation under the model $f_{\theta_1}$ generalize to changes in the parameters $\theta$ while the risk of both classifiers stays approximately the same, i.e.\ $R(f_{\theta_1}) \approx R(f_{\theta_2})$? (b) Will the initially generated counterfactual recommendation under the hypothesis class $\mathcal{F}$ (e.g. regularized linear models) generalize to a model $g$ from a different hypothesis class (e.g. random forest) while $R(f) \approx R(g)$?
For the experiment described in (a), we do not transfer the counterfactual recommendation to any model, but only to those from the $\epsilon$-level set. For the experiments in (b) we extend the below definition to models outside the hypothesis class $\mathcal{F}$.
\begin{definition}[$\epsilon$-level set \citep{marx2019predictive}] 
Given any classifier $f$ and a hypothesis class $\mathcal{F}$, the $\epsilon$-level set around $f$ is the set of all models $g \in \mathcal{F}$ that make at most $\hat{R}(f) + \epsilon$ mistakes over the training data.
\end{definition}
We choose $\epsilon= +/-0.05$. To generate the models from the $\epsilon$-level set we use the \texttt{cv grid search} method from \texttt{scikit learn}. We then use models $g$ within the set and check whether the counterfactual recommendations generated based on $f$ are equally valid under $g$. Particularly, we check two different hypothesis classes: $\mathcal{F}_{Linear}$ and $\mathcal{F}_{Random Forest} = \mathcal{F}_{RF}$. In practice we generate $\tilde{x}(f):= E(x;f)$ and compute $\mathcal{T} = 1/n_{E} \cdot \mathbb{I} [f(\tilde{x}(f)) = g(\tilde{x}(f))]$, where $n_{E}$ is the number of individuals for which counterfactual recommendations are computed and $\mathbb{I}(\cdot)$ denotes the indicator function.

\begin{figure*}
    \centering
\subcaptionbox{HELOC ($f_{Linear} \xrightarrow{} g$).
    \label{fig:transfer_heloc_count}}{\includegraphics[width=0.49\textwidth]{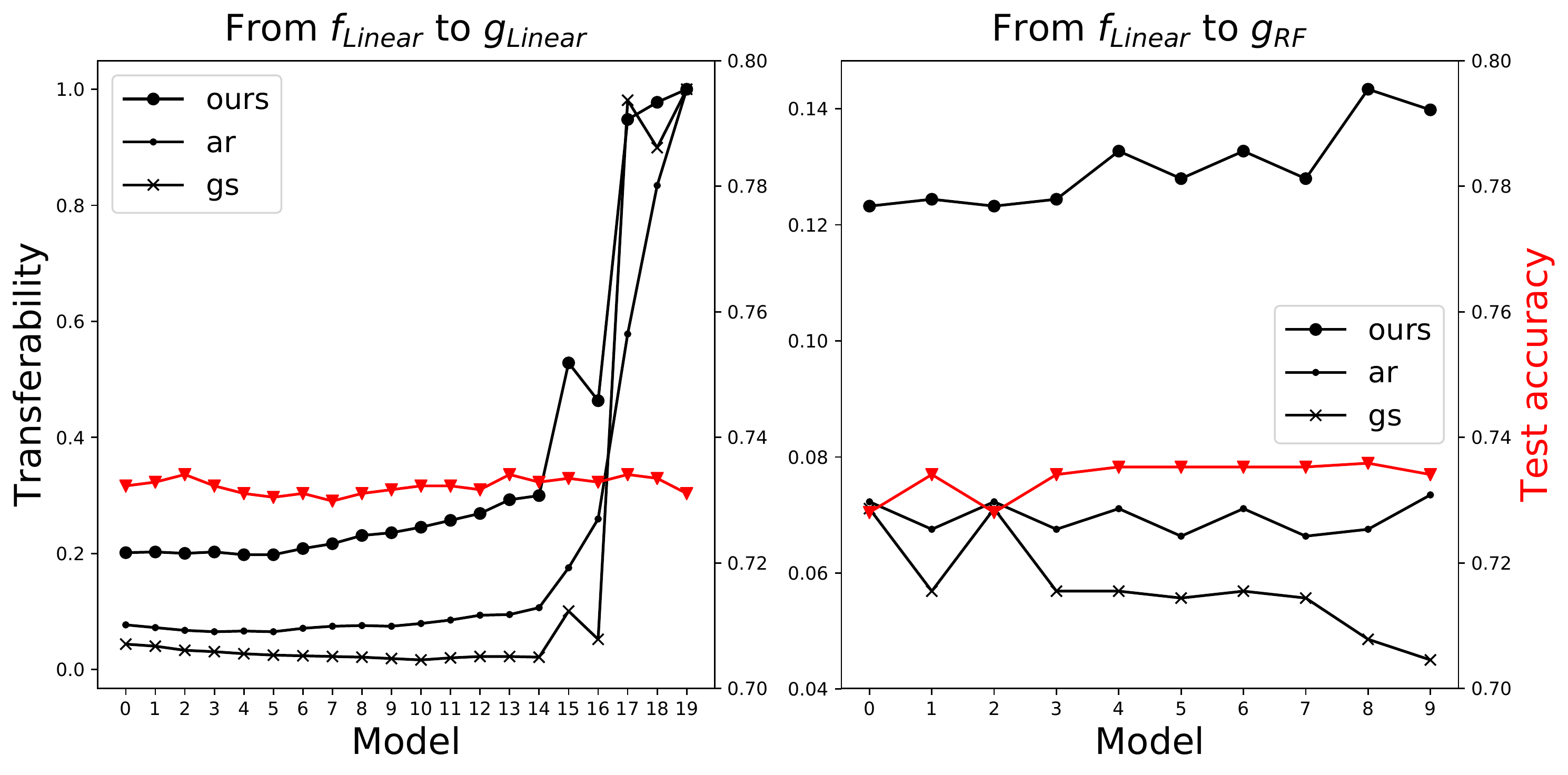}}%
\subcaptionbox{'Give Me Some Credit' ($f_{Linear} \xrightarrow{} g$). \label{fig:transfer_giveme}}{\includegraphics[width=0.49\textwidth]{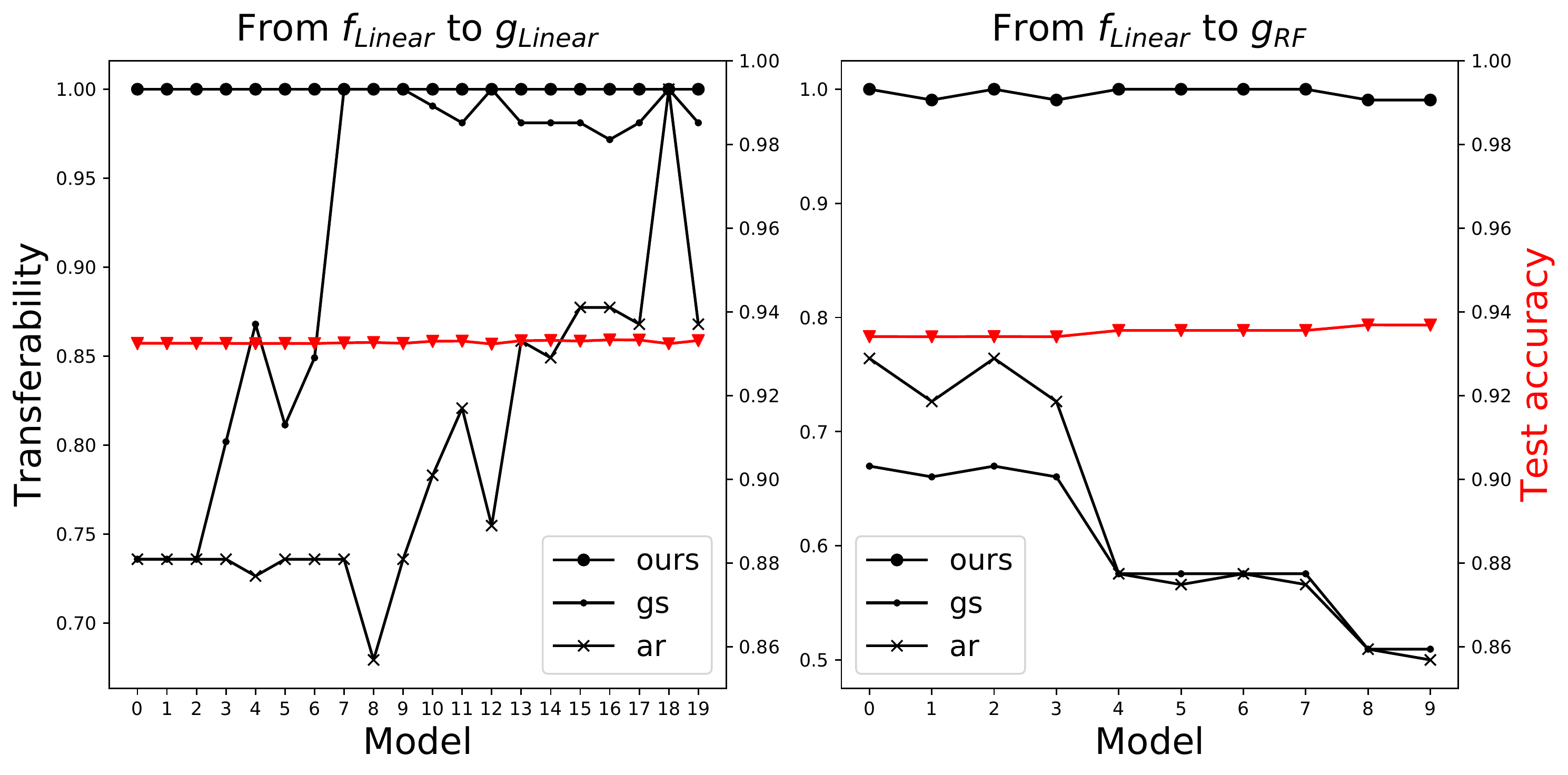}} \vfill 
\subcaptionbox{'Give Me Some Credit' ($f_{RF} \xrightarrow{} g$).\label{fig:transfer_nonlinear_giveme}}{\includegraphics[width=0.49\textwidth]{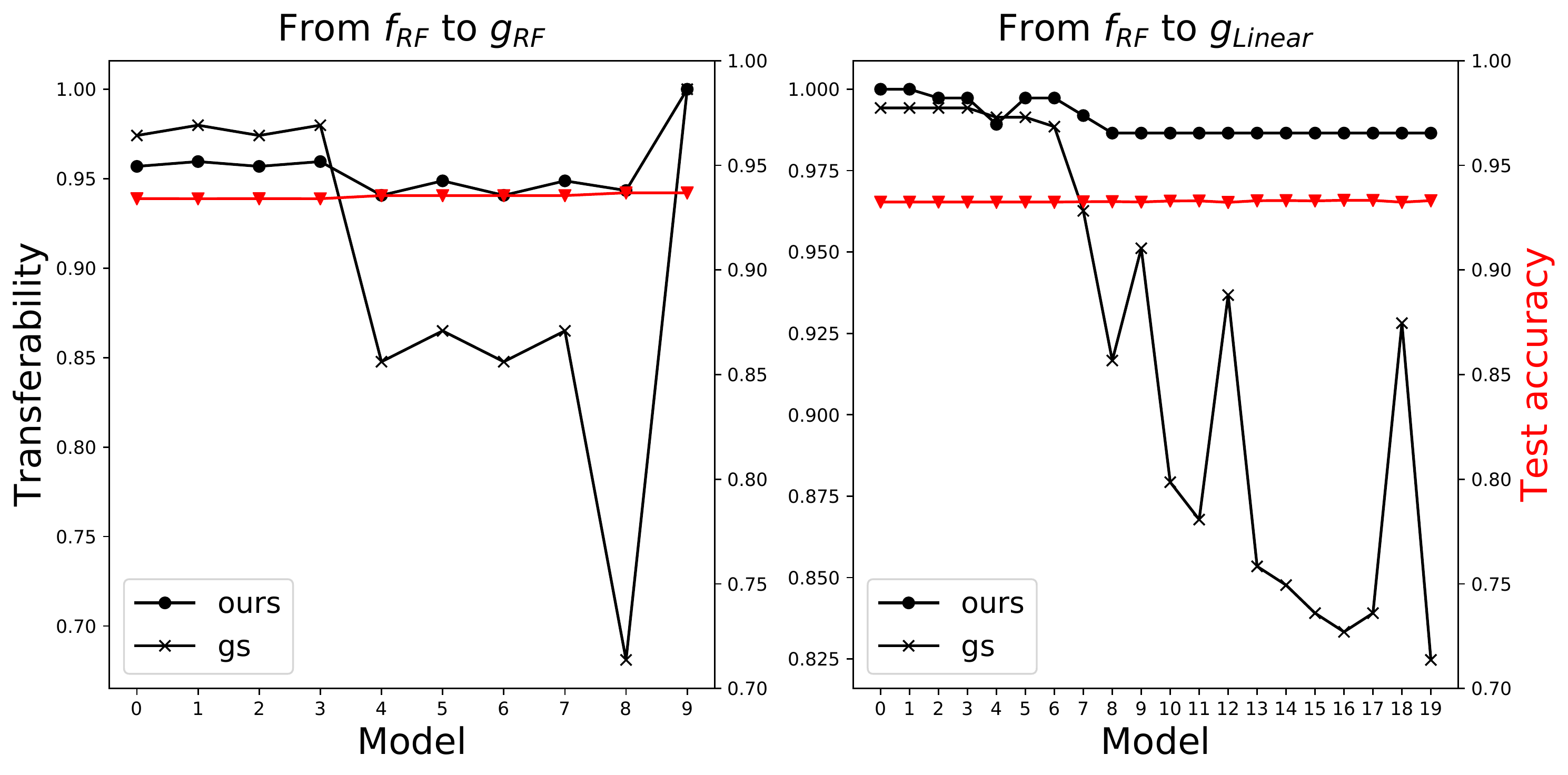}} %
\subcaptionbox{HELOC ($f_{RF} \xrightarrow{} g$).\label{fig:transfer_nonlinear_heloc}}{\includegraphics[width=0.49\textwidth]{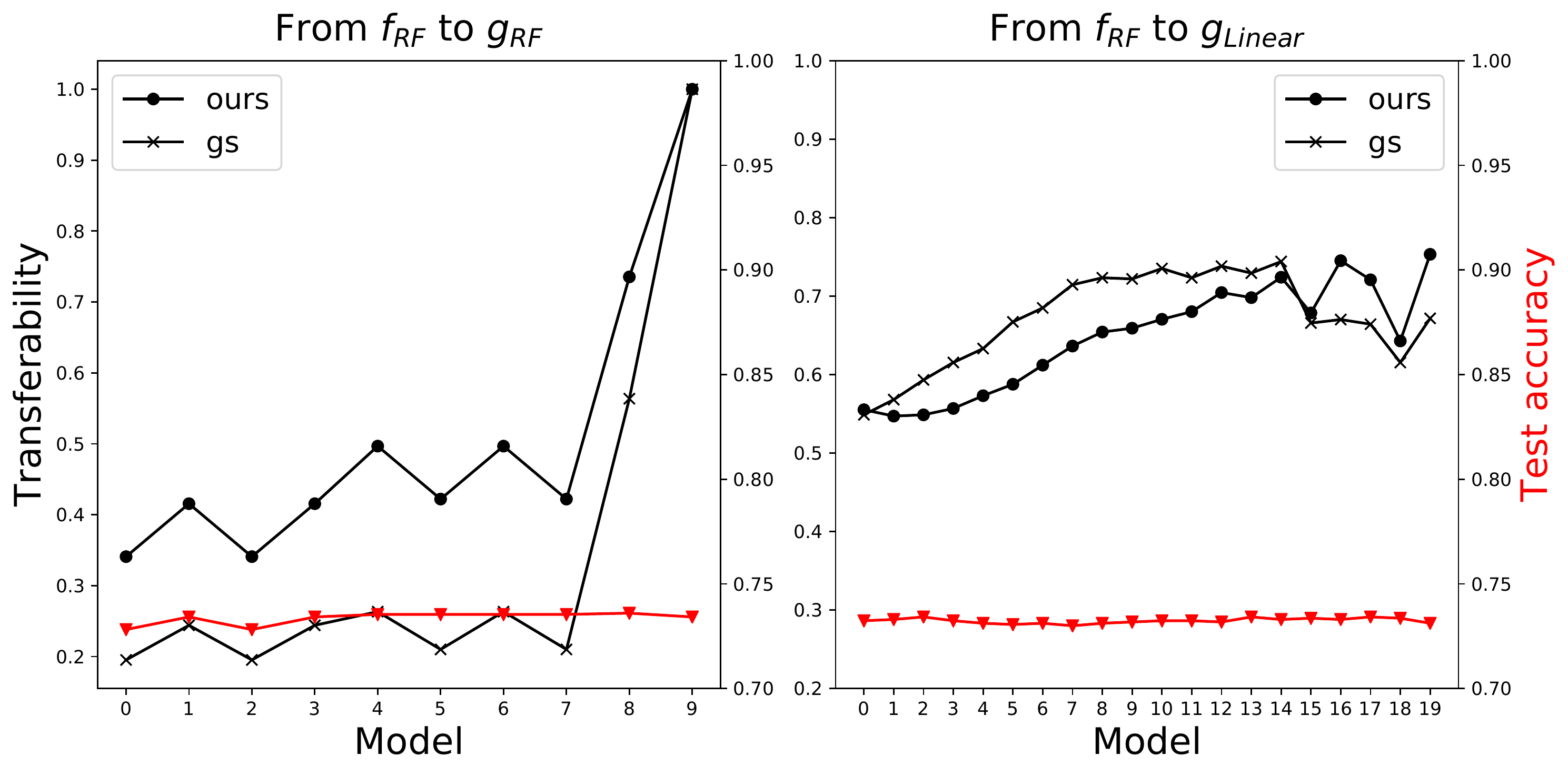}}%
\caption{\textbf{Invariance to predictive multiplicity}. We generate counterfactual explanations according to model $f$ and then check whether they are still valid under model $g$. Left axis: \textcolor{gray}{percentage of counterfactual explanations that are robust to model changes}. Right axis: \textcolor{red}{model accuracy on hold-out test set}. Generally we observe that counterfactuals from OURS are more invariant to model changes than those from GS and AS.}
    \label{fig:transferability}
\end{figure*}

Next, the results are shown in figure \ref{fig:transferability}. The left y-axis depicts how many counterfactual explanations were transferable from model $f$ to model $g$ (It measures $\mathcal{T}$.). The x-axis indicates the model number, and the right y-axis (red graph) shows the model's corresponding test accuracy. The models are usually ordered (in the left column) so that the rightmost model corresponds to the model we used to generate the counterfactual recommendations in the first place. To summarize the results, we would like to stress several points.
\begin{itemize}
    \item [(a)] \textbf{Importance of invariant explanations}. This exercise underlines the importance to learn counterfactual recommendations that are invariant to small (within hypothesis class) and large (between hypothesis class) model perturbations. This is important since the effect of predictive multiplicity makes the model with respect to which we generate counterfactual recommendations look almost arbitrary.
    \item[(b)] \textbf{Data supported counterfactuals are more often model invariant}. The OURS model generates the most robust recommendations: it outperforms GS and AR on all tasks and almost all classifiers. It performs a little worse on the HELOC data set when transferring from $\mathcal{F}_{RF}$ to $\mathcal{F}_{Linear}$ (right panel in figure \ref{fig:transfer_nonlinear_heloc}).
\end{itemize}
In this section, we have empirically investigated whether counterfactual recommendations generalize across models and are thus robust to predictive multiplicity. We found that the data support based methods had superior generalization capabilities. Proposition \ref{proposition:oracle_cost} and \ref{proposition:expected_cost_multiplicity} suggest that these recommendations should also be more costly. We investigate this shortly in the following section.

\subsection{Costs of Counterfactual Recommendations}\label{sec:cost_recommendations}
In order to evaluate the cost of counterfactual suggestions across different models, we use the following two measures \citep{pawelczyk2019learning}:
\begin{align}
    cost_{1}(\tilde{x}^{};x) &= \sum_j |(Q_j(\tilde{x}_j^{})-{Q_j(x_j)}|, \label{eq:total_shift} \\
    cost_{2}(\tilde{x}^{};x) &= \underset{j}{\max}~ |Q_j(\tilde{x}_j^{})-Q_j(x_j)|, \label{eq:max_shift}
\end{align}   
where the subscript denotes the $j$-th component of $x$. The total percentile shift in \eqref{eq:total_shift} can be thought of as a baseline measure for how attainable a certain counterfactual suggestion might be. The maximum percentile shift (MS) in \eqref{eq:max_shift} across all free features reflects the maximum difficulty across all inputs that are subject to change. 

\begin{figure}[htb!]
\centering
\subcaptionbox{HELOC ($f_{Linear}$).\label{fig:percentiles_heloc}}{\includegraphics[width=\columnwidth]{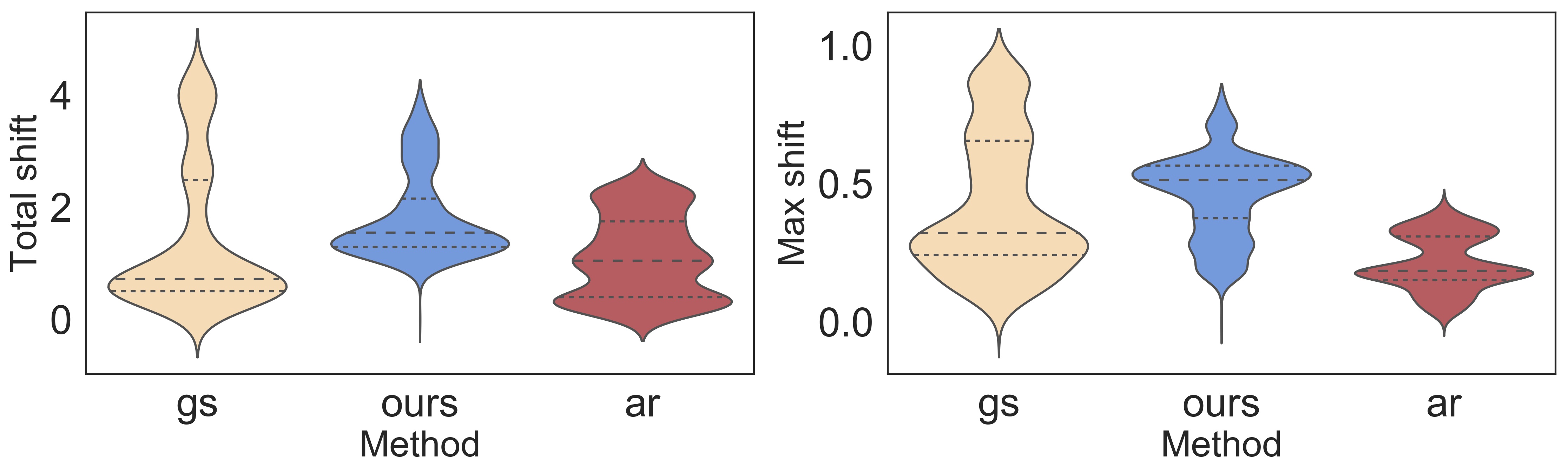}} \vfill
\subcaptionbox{'Give Me Some Credit' ($f_{Linear}$).\label{fig:percentiles_giveme}}{\includegraphics[width=\columnwidth]{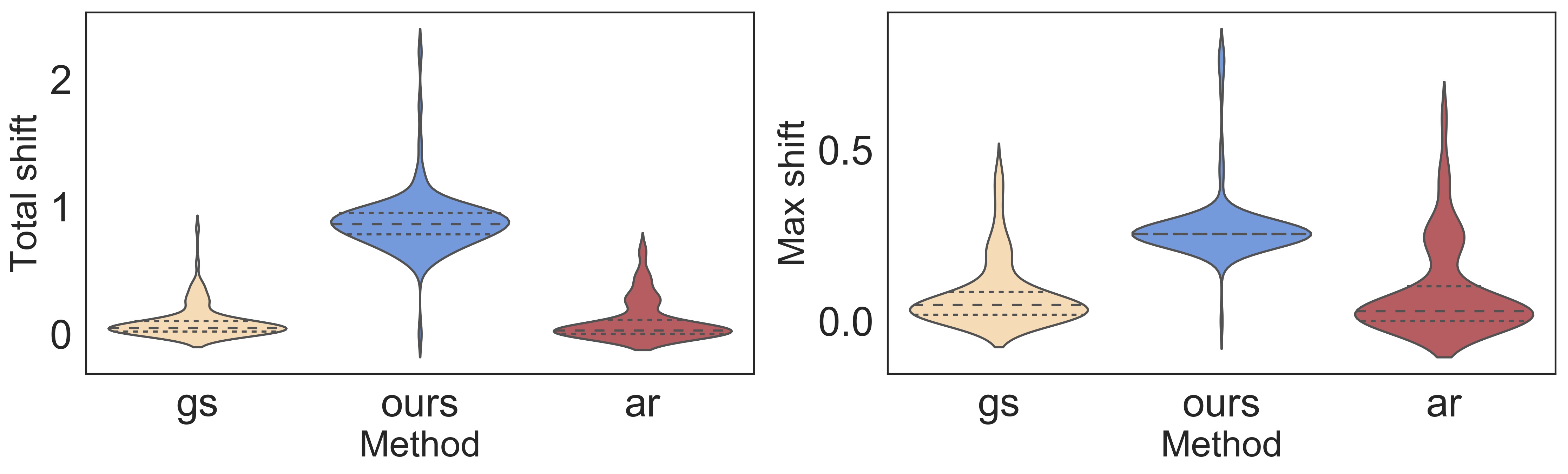}}
\caption{\textbf{Costs of counterfactual recommendations}. The data density based method OURS (it uses a VAE) generates more costly counterfactual recommendations than GS and AR.}
\label{fig:percentiles_all}
\end{figure}

Figure \ref{fig:percentiles_all} shows the resulting plots. The left panel shows violinplots for the distribution of total percentile shifts and the right panel shows these plots for the maximum percentile shift. From the plots it becomes clear that the OURS method generates counterfactual recommendations that tend to have both higher total and maximum percentile shifts. This holds for both data sets.

In the next section, we briefly investigate why OURS works better in producing invariant recommendations. We will also understand why it generates higher costs.

\subsection{Why Data Supported Counterfactuals tend to Generate more Invariant Explanations}
\paragraph{\textbf{On robustness}.} 
\begin{figure}[htb!]
\centering
\includegraphics[width=0.95\columnwidth]{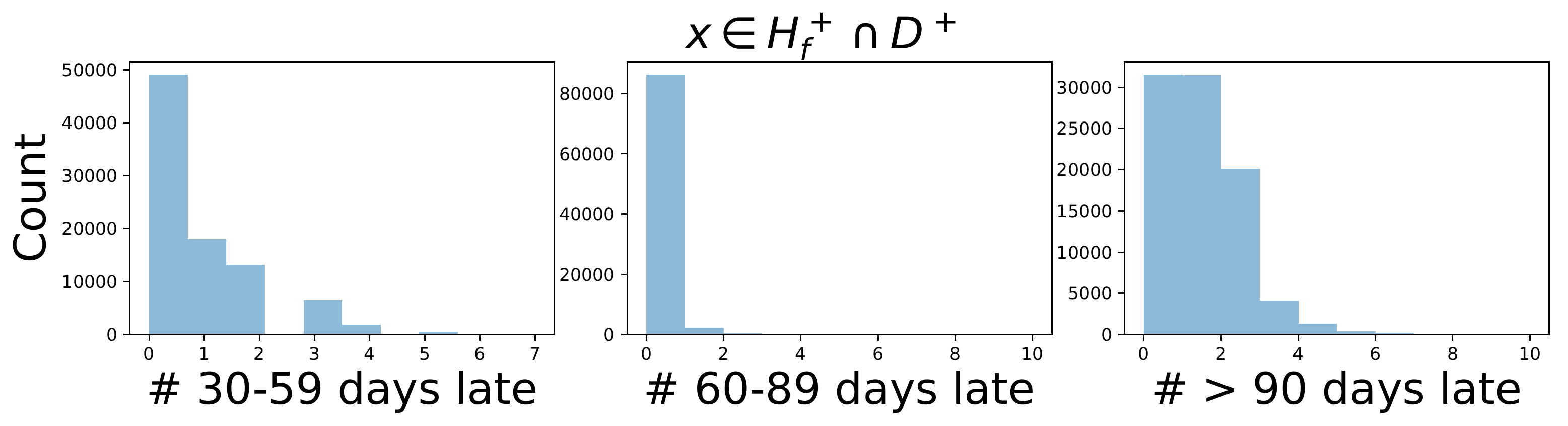} \vfill
\includegraphics[width=0.95\columnwidth]{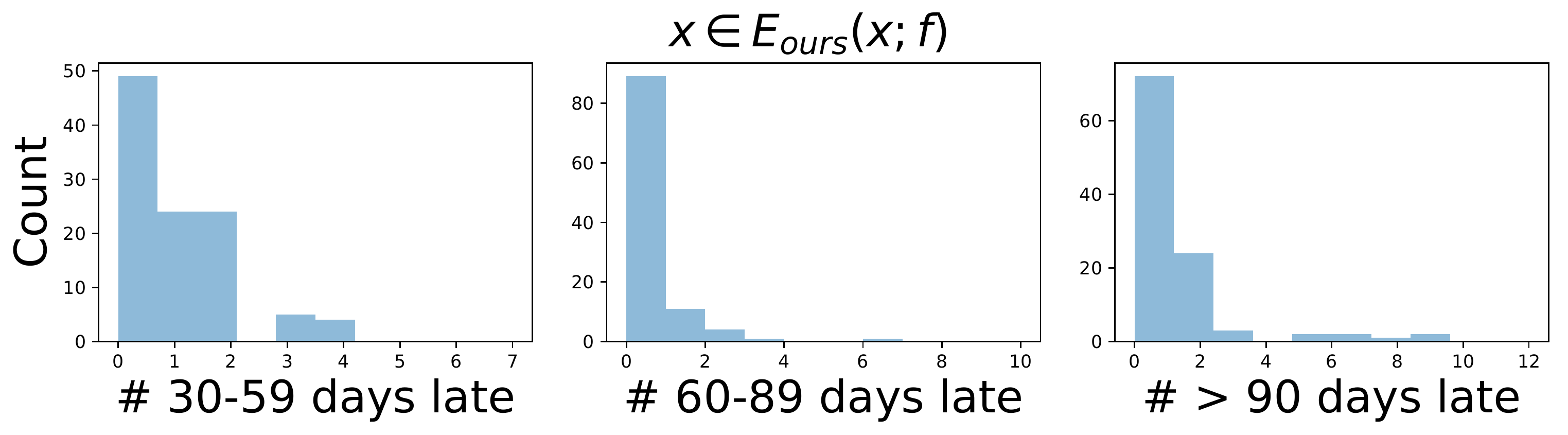} \vfill
\includegraphics[width=0.95\columnwidth]{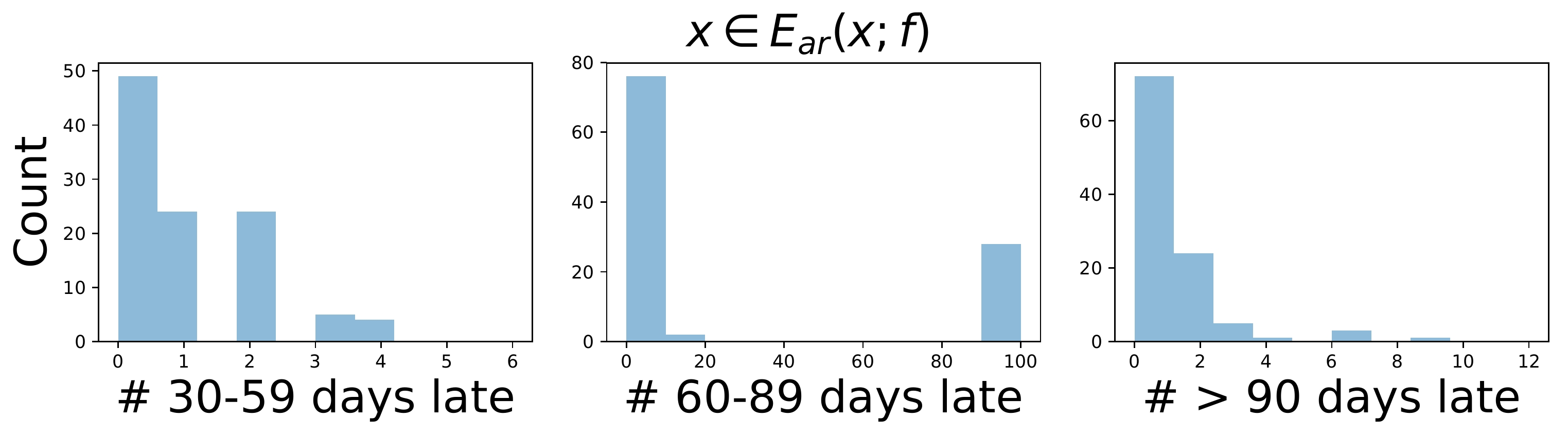}
\caption{\textbf{Timeliness and counterfactual suggestions} for the "Give Me Some Credit" data. (Top row) Histogram of three inputs related to timeliness of loan payments for individuals from $H_f^+ \cap D^+.$ (Middle \& Bottom row) Histogram for counterfactual recommendations on the test set for explanations from OURS (middle) and AR (bottom).}
\label{fig:histogram_timeliness}
\end{figure}

Next, we zoom into one particular set of inputs. We look at the distribution of three inputs: \texttt{30-59 days late}, \texttt{60-89 days late}, \texttt{>89 days late}, which one could summarize as \emph{timeliness} of individuals' payments. The first row in figure \ref{fig:histogram_timeliness} shows the distribution of the three inputs for which we have that $H_f^+ \cap D^+$: in words, correctly classified individuals make their loan payments on time. The counterfactual recommendations generated by OURS, $E_{ours}(\bm{x};f)$, second row of figure \ref{fig:histogram_timeliness}, follow this distribution quite closely. The last row shows the distribution induced by $E_{ar}(\bm{x};f)$, which is not close to the one of the correctly classified individuals for \texttt{\# 60-89 days late}.

\paragraph{\textbf{On costs}.}
Recall from section \ref{sec:cost_recommendations} and figure \ref{fig:percentiles_all} that OURS' recommendations are more costly. Now, let us consider figure \ref{fig:histogram_timeliness} again. To generate lower cost recommendations, AR needs to be very close to the original inputs. Thus, it often suggests to leave the timeliness inputs unchanged (third row in figure \ref{fig:histogram_timeliness}) or only change a subset of them. This, however, appears counter intuitive (we discuss this issue further in appendix \ref{appendix:further_evaluations}).

\section{CONCLUSION}\label{sec:discussion}
In light of the fact that counterfactual recommendations can have an huge impact on individuals' lives, there existed remarkably little work regarding cost guarantees for existing methods. In this work, we have taken a step towards filling this void. We theoretically analyzed the cost of counterfactual recommendations for \emph{sparse} and \emph{data supported} counterfactual recommendations. Most notably, we obtained the following insights: first, \emph{data supported} counterfactual recommendations are at least as costly as \emph{sparse} ones. Second, if assumption \textbf{A1} (classifier is stable) is violated, the cost of counterfactual recommendations under model multiplicity can be substantially higher than under one fixed model. 
Therefore, counterfactual recommendations are ideally based on (explanation) models that causally (and thus invariantly) relate inputs to targets to avoid the impact of predictive multiplicity on counterfactuals.

Our results have thus guided us to an interesting question for future research: \emph{can one generate invariant counterfactual recommendations with minimal costs}?

To establish trustworthy (semi-) automated ML systems with humans in the loop, it is crucial to provide counterfactual recommendations with cost guarantees, which humans can rely on when working towards their goals. Therefore, we hope that our work can help practitioners make more informed decisions on which type of recommendation method to choose in the future.




\clearpage
\bibliographystyle{plainnat}
\bibliography{example_paper.bib}


\onecolumn
\title{Appendix: Challenging common practices for the generation of counterfactual explanations}
\appendix
\maketitle
\begin{appendices}

\subsubsection*{Acknowledgements}
We would like to thank Hamed Jalali, Charlie Marx and the anonymous reviewers for insightful comments and suggestions.

\section{Further experimental evaluations}\label{appendix:further_evaluations}
\paragraph{\textbf{On semantics}.}
Consider table \ref{tab:recommendation}. While GS' recommendations tend to have low costs, they often take on ambiguous values. We have marked the critical values in \textcolor{red}{red}. AR's recommendations tend to make sense, if one inspects them input value by input value. However, often they run into logical inconsistencies, which we highlighted in \textcolor{blue}{blue} in table \ref{tab:recommendation}. So does it make sense to tell someone to be more often on time for the 30-59 days range while demanding that the person should be paying more often 60-89 days late? Probably not. The \emph{data support} counterfactual recommendations from OURS, in contrast, appear to make sense and seem consistent. There exist multiple of these examples in the generated explanations and many of them follow a similar pattern. \emph{Most importantly, this demonstrates that end users could find it troublesome to comprehend what causes a classifier to behave a certain way, versus what causes the world to behave in a certain way}.

\begin{table*}[htb!]
\begin{tabular}{lrrrrrrrr}
\toprule
\multirow{2}{*}{Model} & \multicolumn{8}{c}{set of mutable inputs} \\
\cmidrule(l){2-9}
  &  rev.util. &  \#30-59 d. late &  debt ratio &   income &  \# credit &  $>$ 90 d. late &  \# r. est. loans &  \# 60-89 d. l. \\
\midrule
$x \in H_f^- $  &   1.00 &          3.00 &    0.19 &  2700.00 &          3.00 &              4.00 &              0.00 &         0.00 \\
\cmidrule(l){1-9}
GS &   1.12 &          \textcolor{red}{2.77} &    0.24 &  2699.92 &          \textcolor{red}{3.03} &              \textcolor{red}{4.08} &             \textcolor{red}{-0.13} &         \textcolor{red}{0.25} \\
AR &   1.00 &          \textcolor{blue}{2.00} &    0.19 &  2700.00 &          3.00 &              4.00&              0.00 &         \textcolor{blue}{2.00} \\
OURS  &   0.97&          0.00 &    0.18 &  2753.82 &          3.00 &              0.00&              0.00 &         0.00 \\
\bottomrule
\end{tabular}
    \caption{\textbf{Illustrative example, comparing semantics of recommendations from GS, OURS and AR}. The instance $x \sim p_{data}$ was negatively classified by the prediction model $f$. For this individual, the immutable inputs are fixed at age = 36  and \# dependents = 3. \textcolor{red}{red}: ambiguous values. \textcolor{blue}{blue}: inconsistent values. }
    \label{tab:recommendation}
\end{table*}

\begin{figure*}[htb!]
\centering
\begin{subfigure}{\textwidth}
\includegraphics[width=\textwidth]{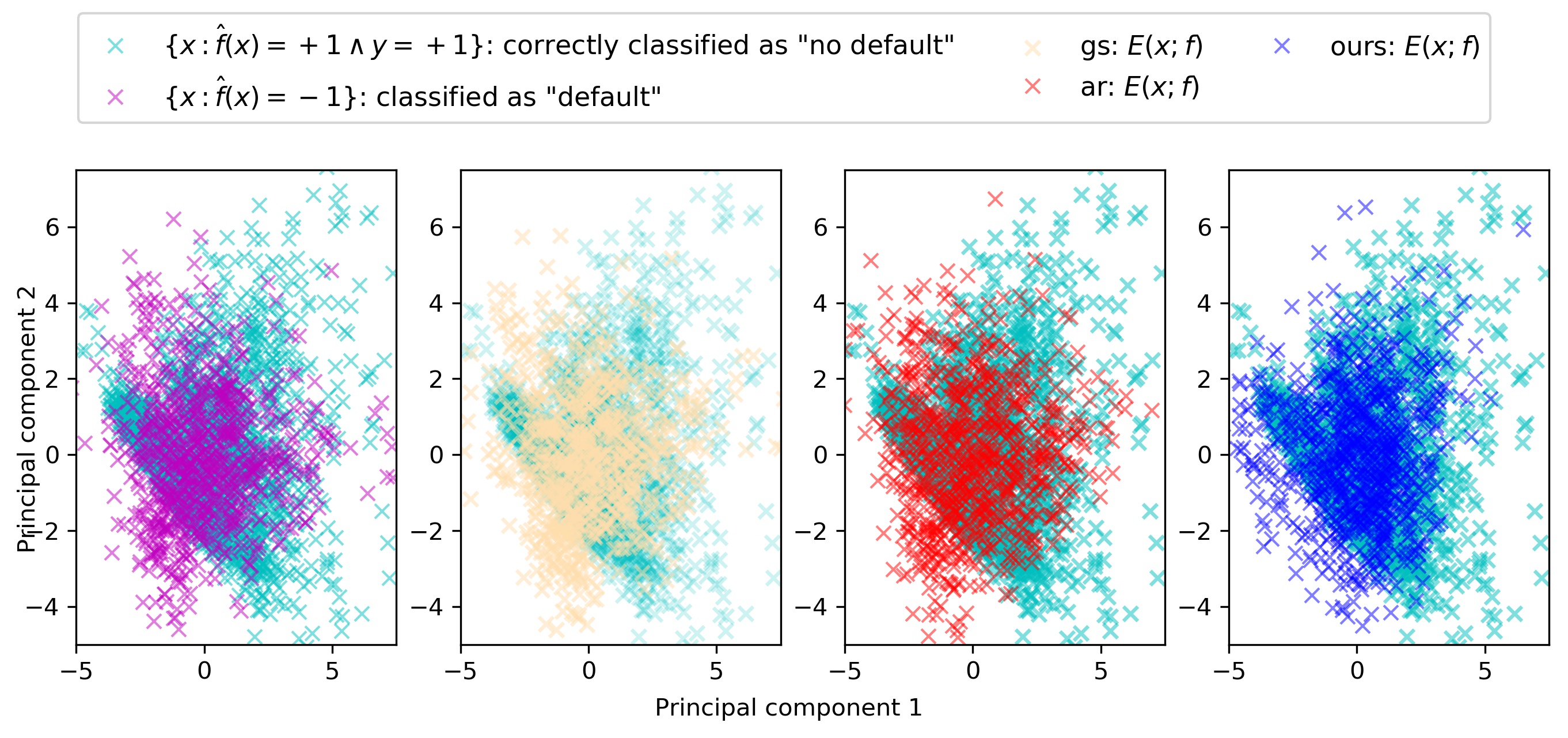}
\caption{HELOC}
\end{subfigure} \vfill
\begin{subfigure}{\textwidth}
\includegraphics[width=\textwidth]{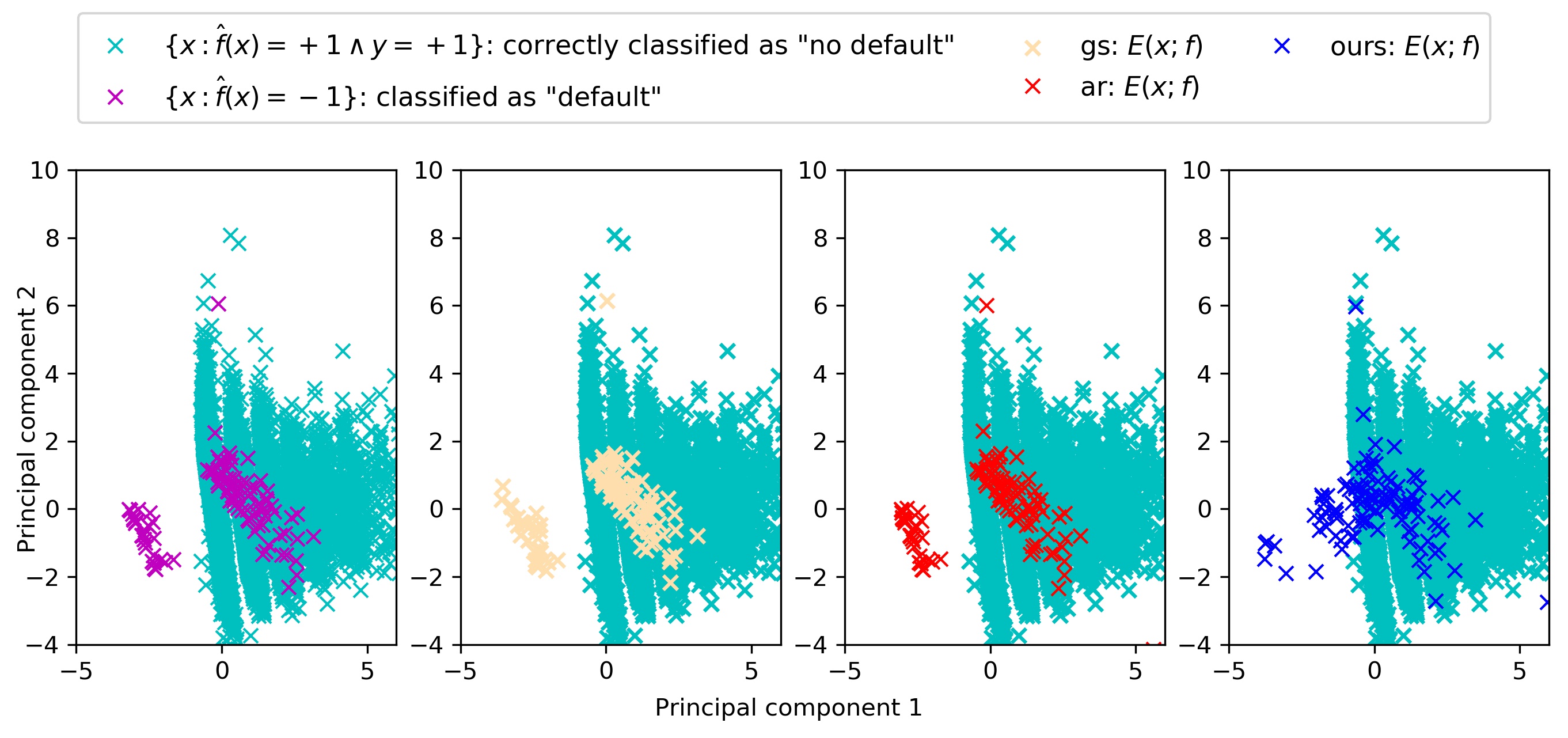}
\caption{Give Me Some Credit}
\end{subfigure}
\caption{\textbf{First and second principal components} of $H_f^+ \cap D^+$ (cyan), $H^-$ (magenta) and counterfactual recommendations $E(\bm{x}; f)$, where $f$ denotes the pretrained regularized linear regression classifier (in this case). Recall that $f(E(\bm{x}; f)) = +1$. AR's (red) and GS' (yellow) latent space representation of the generated counterfactual recommendations remain very close to the incorrectly classified representation (purple). OURS (blue, right most) rotates and pushes the latent space closer to the one of the correctly classified observations $H_f^+ \cap D^+$ (cyan).}
\label{fig:pca}
\end{figure*}

\paragraph{\textbf{More on robustness}.} 
Recall that we wish to find recommendations for the negative predicted individuals, $H_f^-$. As opposed to the other methods, the OURS method \emph{pushes the negative predicted individuals towards data points from the correctly classified individuals}, $H_f^+ \cap D^+$. To show this for all explanations, we compute the first two principal components of $E_{ar}(\bm{x};f)$, $E_{gs}(\bm{x};f)$ and $E_{ours}(\bm{x};f)$ and compare them to $H_f^-$, $H_f^+ \cap D^+$ (see figure \ref{fig:pca}).


\section{Data and Implementations}\label{sec:appendix_data}
\subsection{Real world example: ``\texttt{Give Me Some Credit}''}
In the following, we list the specified pretrained classification models as well as the parameter specification used for the experiments. We use 80 percent of the data as our training set and the remaining part is used as the holdout test set. Additionally, we allow $f$ and $g$ access to all features, i.e.. to the mutable and immutable ones. The state of features can be found in table \ref{tab:state_givme}.

\paragraph{\textbf{AR} \citep{ustun2019actionable}.}
The AR algorithm requires to choose both an action set and free and immutable features. The implementation can be found here: \url{https://github.com/ustunb/actionable-recourse}. We specify that the \emph{DebtRatio} feature can only move downward \citep{ustun2019actionable}. The AR implementation has a default decision boundary at $0$ and therefore one needs to shift the boundary. We choose $p_{AR} = 0.50$, adjusting the boundary appropriately. Finally, we set the linear programming optimizer to \emph{cbc}, which is based on an open-access \texttt{python} implementation.

\paragraph{\textbf{GS} \citep{laugel2017inverse}.}
GS is based on a version of the YPHL algorithm. As such we have to choose appropriate step sizes in our implementation to generate new observations from the sphere around $\bm{x}$. We choose a step size of 0.1.

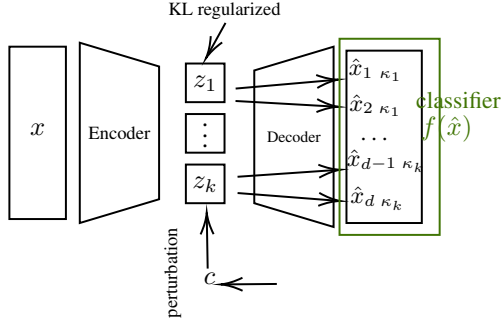
\begin{figure}
\centering
\tikzset{every picture/.style={line width=0.75pt}} 
\begin{tikzpicture}[x=0.75pt,y=0.75pt,yscale=-1,xscale=1]

\draw   (127,104) -- (167,116) -- (167,183) -- (127,195) -- cycle ;
\draw   (120,106) -- (120,193) -- (91.5,193) -- (91.5,106) -- cycle ;
\draw   (180.5,114.5) -- (200,114.5) -- (200,134) -- (180.5,134) -- cycle ;
\draw   (180.5,140) -- (200,140) -- (200,159.5) -- (180.5,159.5) -- cycle ;
\draw   (180.5,165.5) -- (200,165.5) -- (200,185) -- (180.5,185) -- cycle ;
\draw   (255,197) -- (215,185) -- (215,118) -- (255,106) -- cycle ;
\draw   (299.5,108) -- (299.5,195) -- (261.5,195) -- (261.5,108) -- cycle ;
\draw    (205.6,127.14) -- (255.41,122.8) ;
\draw [shift={(257.4,122.62)}, rotate = 535.02] [color={rgb, 255:red, 0; green, 0; blue, 0 }  ][line width=0.75]    (10.93,-3.29) .. controls (6.95,-1.4) and (3.31,-0.3) .. (0,0) .. controls (3.31,0.3) and (6.95,1.4) .. (10.93,3.29)   ;

\draw    (191.5,218) -- (191.05,191.76) ;
\draw [shift={(191.02,189.76)}, rotate = 449.03] [color={rgb, 255:red, 0; green, 0; blue, 0 }  ][line width=0.75]    (10.93,-3.29) .. controls (6.95,-1.4) and (3.31,-0.3) .. (0,0) .. controls (3.31,0.3) and (6.95,1.4) .. (10.93,3.29)   ;

\draw    (226.38,225.76) -- (200.14,225.98) ;
\draw [shift={(198.14,226)}, rotate = 359.52] [color={rgb, 255:red, 0; green, 0; blue, 0 }  ][line width=0.75]    (10.93,-3.29) .. controls (6.95,-1.4) and (3.31,-0.3) .. (0,0) .. controls (3.31,0.3) and (6.95,1.4) .. (10.93,3.29)   ;

\draw    (205.54,133.41) -- (255.46,136.24) ;
\draw [shift={(257.46,136.35)}, rotate = 183.24] [color={rgb, 255:red, 0; green, 0; blue, 0 }  ][line width=0.75]    (10.93,-3.29) .. controls (6.95,-1.4) and (3.31,-0.3) .. (0,0) .. controls (3.31,0.3) and (6.95,1.4) .. (10.93,3.29)   ;

\draw    (206.56,171.66) -- (256.45,168.23) ;
\draw [shift={(258.44,168.1)}, rotate = 536.0699999999999] [color={rgb, 255:red, 0; green, 0; blue, 0 }  ][line width=0.75]    (10.93,-3.29) .. controls (6.95,-1.4) and (3.31,-0.3) .. (0,0) .. controls (3.31,0.3) and (6.95,1.4) .. (10.93,3.29)   ;

\draw    (206.58,179.87) -- (256.43,183.74) ;
\draw [shift={(258.42,183.89)}, rotate = 184.44] [color={rgb, 255:red, 0; green, 0; blue, 0 }  ][line width=0.75]    (10.93,-3.29) .. controls (6.95,-1.4) and (3.31,-0.3) .. (0,0) .. controls (3.31,0.3) and (6.95,1.4) .. (10.93,3.29)   ;

\draw    (200.5,94) -- (191.6,110.08) ;
\draw [shift={(190.64,111.83)}, rotate = 298.95] [color={rgb, 255:red, 0; green, 0; blue, 0 }  ][line width=0.75]    (10.93,-3.29) .. controls (6.95,-1.4) and (3.31,-0.3) .. (0,0) .. controls (3.31,0.3) and (6.95,1.4) .. (10.93,3.29)   ;

\draw  [color={rgb, 255:red, 65; green, 117; blue, 5 }  ,draw opacity=1 ] (308.5,102) -- (308.5,201) -- (258,201) -- (258,102) -- cycle ;

\draw (147,149.5) node  [font=\footnotesize] [align=left] {{\scriptsize Encoder}};
\draw (235,151.5) node  [font=\tiny] [align=left] {Decoder};
\draw (105.75,147.5) node    {$x$};
\draw (190.25,124.25) node    {$z_{1}$};
\draw (190.25,175.25) node    {$z_{k}$};
\draw (190.25,146.75) node    {$\vdots $};
\draw (277.75,117.5) node  [font=\footnotesize,rotate=-359.76]  {$\hat{x}_{1_{\ } \kappa_{1}} \ $};
\draw (277.75,135.5) node  [font=\footnotesize,rotate=-359.76]  {$\hat{x}_{2\ \kappa_{1}} \ $};
\draw (278.75,182.5) node  [font=\footnotesize,rotate=-359.76]  {$\hat{x}_{d\ \kappa_{k}} \ $};
\draw (284.75,163.5) node  [font=\footnotesize,rotate=-359.76]  {$\hat{x}_{d-1\ \kappa_{k} \ } \ $};
\draw (279.5,152.5) node  [font=\small,rotate=-89.78]  {$\vdots $};
\draw (200,88) node  [font=\scriptsize] [align=left] {KL regularized};
\draw (312,147) node  [color={rgb, 255:red, 65; green, 117; blue, 5 }  ,opacity=1 ]  {$f(\hat{x})$};
\draw (318,133) node  [font=\footnotesize,color={rgb, 255:red, 65; green, 117; blue, 5 }  ,opacity=1 ] [align=left] {classifier};
\draw (192.5,223) node    {$c$};
\draw (174,220) node  [font=\scriptsize,rotate=-270] [align=left] {perturbation};

\end{tikzpicture}
\caption{Schematic figure for counterfactual search from the OURS model \citep{pawelczyk2019learning}. The latent representation ideally learns independent concepts denoted by $\kappa_1, ..., \kappa_k$ (e.g. timeliness, overall financial situation, etc.).}
\label{fig:ours_search}
\end{figure}

\paragraph{\textbf{OURS} \citep{pawelczyk2019learning}.} We used the (H)VAE implemention as described here: \url{https://github.com/probabilistic-learning/HI-VAE} \citep{nazabal2018handling}. Random search in the latent space was conducted to find counterfactual recommendations, using the YPHL algorithm \citep{laugel2017inverse}. We made the following choices. We set the latent space dimension of both $\bm{s}$ and $\bm{z}$ to 3 and 6, respectively. For training, we used 15 epochs. Table \ref{tab:state_givme} gives details about the chosen likelihood model for each input. For count inputs, we use the Poisson likelihood model, while for inputs with a support on the positive part of the real line we choose log normal distributions.
\begin{table}[htb!]
    \centering
     \begin{adjustbox}{max width=\columnwidth}
    \begin{tabular}{lcc}
    \toprule
         Inputs & Mutable & Model \\
         \cmidrule(lr){1-3}
         \emph{Revolving Utilization Of Unsecured Lines} & Y & log Normal \\ 
         \emph{Age} & N & Poisson  \\
         \emph{Number Of Times 30-59 Days Past Due Not Worse} & Y & Poisson \\
         \emph{Debt Ratio} & Y & log Normal \\
         \emph{Monthly Income} & Y & log Normal  \\
         \emph{Number Open Credit Lines And Loans} & Y & Poisson \\
         \emph{Number Of Times 90 days Late} & Y & Poisson  \\
         \emph{Number Real Estate Loans Or Lines} & Y & Poisson \\
         \emph{Number Of Times 60-89 Days Past Due Not Worse} & Y & Poisson  \\
         \emph{Number Of Dependents} & N & Poisson \\
         \bottomrule
    \end{tabular}
    \end{adjustbox}
    \caption{``\texttt{Give Me Some Credit}'': State of inputs and likelihood models.}
    \label{tab:state_givme}
\end{table}

\begin{table}
    \centering
     \begin{adjustbox}{max width=\columnwidth}
    \begin{tabular}{lccc}
    \toprule
         Input & Mutable & Model \\
         \cmidrule(lr){1-3}
         \emph{MSinceOldestTradeOpen} & N & Poisson  \\
         \emph{AverageMInFile} & N & Poisson \\ 
         \emph{NumSatisfactoryTrades} & Y & Poisson \\
         \emph{NumTrades60Ever/DerogPubRec} & Y & Poisson\\
         \emph{NumTrades90Ever/DerogPubRec} & Y & Poisson \\
         \emph{NumTotalTrades} & Y & Poisson \\
         \emph{PercentInstallTrades} & Y & Poisson \\
         \emph{MSinceMostRecentInqexcl7days} & Y & Poisson \\
         \emph{NumInqLast6M} & Y & Poisson \\
         \emph{NetFractionRevolvingBurden} & Y & Poisson \\
         \emph{NumRevolvingTradesWBalance} & Y & Poisson \\
         \emph{NumBank/NatlTradesWHighUtilization} & Y & Poisson \\
         \emph{ExternalRiskEstimate} & N & Poisson  \\
          \emph{MPercentTradesNeverDelq} & Y & Poisson \\
          \emph{MaxDelq2PublicRecLast12M} & Y & Poisson  \\
          \emph{MaxDelqEver} & Y & Poisson\\
          \emph{NumTradesOpeninLast12M} & Y & Poisson\\
          \emph{NumInqLast6Mexcl7days} & Y & Poisson  \\
          \emph{NetFractionRevolvingBurden} & Y & Poisson \\
          \emph{NumInstallTradesWBalance} & Y & Poisson \\
          \emph{NumBank2NatlTradesWHighUtilization} & Y & Poisson  \\
          \emph{PercentTradesWBalance} & Y & Poisson \\
         \bottomrule
    \end{tabular}
    \end{adjustbox}
    \caption{\texttt{HELOC}: State of inputs and likelihood models.}
    \label{tab:state_default}
\end{table}

\subsection{Real world example: \texttt{HELOC}}


The \emph{Home Equity Line of Credit (HELOC)} data set consists of credit applications made by homeowners in the US, which can be obtained from the FICO community.\footnote{\url{https://community.fico.com/s/explainable-machine-learning-challenge?tabset-3158a=2}.} The task is to use the applicant's information within the credit report to predict whether they will repay the HELOC account within 2 years. Table \ref{tab:state_default} gives an overview of the available inputs and the corresponding assumed likelihood models.

\paragraph{\textbf{AR and GS}} As before. Additionally, we do not specify how features have to move.
\paragraph{\textbf{OURS}}
We set the latent space dimension of both $\bm{s}$ and $\bm{z}$ to 12 and 10, respectively. For training, we used 60 epochs. Table \ref{tab:state_default} gives details about the chosen likelihood model for each feature. The rest remains as before.

\section{Proof of proposition \ref{proposition:oracle_cost}}\label{appendix:oracle_cost}
\begin{proof}
Let us consider an $\bm{x}_1 \in H_f^+$, i.e. $f(\bm{x}_1) = +1 = f(h(\tilde{\bm{z}}))$. By the assumption of the generative model in the main text, we know that $h(\bm{z})= \bm{x}$. We have $\bm{c}_{D}(\tilde{z}) = \lVert \bm{x} - h(\tilde{\bm{z}})\rVert$ = $\lVert (\bm{x} - \bm{x}_1) + (\bm{x}_1- h(\tilde{\bm{z}}))\rVert \leq \lVert \bm{x} - \bm{x}_1 \rVert + \lVert \bm{x}_1- h(\tilde{\bm{z}}) \rVert$, where we used the triangle inequality. By (1), we have that $\lVert \bm{x}_1- h(\tilde{\bm{z}}) \rVert \leq \lVert \bm{x}_1- h(\bm{z}) \rVert$. Hence, we can write  $\bm{c}_{D}(\tilde{z}) \leq 2\lVert \bm{x}_1- h(\bm{z})\rVert = 2 \lVert \bm{x}_1- \bm{x}\rVert$. Now, minimizing $\bm{x}$ over $\mathcal{E}_S$ (recall definition \ref{def:rec_unrestricted} from the main text) gives the desired result.
\end{proof}

\section{Proof of proposition \ref{proposition:expected_cost_multiplicity}}\label{appendix:expected_cost_multiplicity}
From now on, we suppress the dependence of $f(x):=f$ and $g(x):=g$ on $\bm{x} := x$. For brevity, we sometimes say $A := H_{f}^{-} \cup H_{g}^{-}$ and $\pi_{-} = 1-\pi$. $\pi = Pr_{H_f^- \cap H_g^-}(y=1)$; $\pi_f = Pr_{H_f^-}(y=1)$; $\pi_g = Pr_{H_g^-}(y=1)$

\subsection{Main argument}
\begin{proof}
We first expand the $\overline{cost}(f,g)_{H_{f}^{-} \cup H_{g}^{-}}$.
\begin{align*}
        \overline{cost}(f,g)_{H_{f}^{-} \cup H_{g}^{-}} & = \mathbb{E}_{H_{f}^{-} \cup H_{g}^{-}}[c^*(f,g,x)] \\
    & = \pi \cdot [\mathbb{E}_{A \cap D^{+}}[c^*|f\leq 0 ,g\leq 0] P_{A \cap D^{+}}(f\leq 0 ,g\leq 0) 
     + \mathbb{E}_{A \cap D^{+}}[c^*|f\leq 0 ,g > 0] P_{A \cap D^{+}}(f\leq 0 ,g > 0) \\
    & + \mathbb{E}_{A \cap D^{+}}[c^*|f>0,g>0] P_{A \cap D^{+}}(f>0,g>0)  + \mathbb{E}_{A \cap D^{+}}[c^*|f>0,g\leq 0] P_{A \cap D^{+}}(f>0,g\leq0)] \\ & +\pi_{-} [\mathbb{E}_{A \cap D^{-}}[c^*|f\leq 0 ,g\leq 0] P_{A \cap D^{-}}(f\leq 0 ,g\leq 0)  + \mathbb{E}_{A \cap D^{-}}[c^*|f\leq 0 ,g > 0] P_{A \cap D^{-}}(f\leq 0 ,g > 0) \\
    & + \mathbb{E}_{A \cap D^{-}}[c^*|f>0,g>0] P_{A \cap D^{-}}(f>0,g>0)  + \mathbb{E}_{A \cap D^{-}}[c^*|f>0,g\leq 0] P_{A \cap D^{-}}(f>0,g\leq0)] \\
    \end{align*}
Moreover, note that $\lvert \mathbb{E}_{H_f^- \cap D^+}[f|f\leq 0] \rvert  \leq c^{max}_{H_f^-}(f)$, and $\lvert \mathbb{E}_{H_f^- \cap D^-}[f|f> 0] \rvert  \leq c^{max}_{H_f^-}(f)$. Analogously for the classifier $g$. And hence we can write for the classifier $f$:
\begin{align}
     - \pi_f  P_{H_{f}^{-} \cap D^+}(f\leq0)  \mathbb{E}_{H_{f}^{-} \cap D^+}[f|f\leq 0]
    &= 2  \pi_f  P_{H_{f}^{-} \cap D^+}(f\leq0)  \lvert \mathbb{E}_{H_{f}^{-} \cap D^+}[f|f\leq 0] \rvert \notag \\
    & + \pi_f  P_{H_{f}^{-} \cap D^+}(f\leq0)  \mathbb{E}_{H_{f}^{-} \cap D^+}[f|f\leq 0], \notag \\
    &\leq 2  P_{H_{f}^{-} \cap D^+}(f\leq0)  \pi_f  c^{max}_{H_{f}^{-}}(f) 
    + \pi_f  P_{H_{f}^{-} \cap D^+}(f\leq0)  \mathbb{E}_{H_{f}^{-} \cap D^+}[f|f\leq 0]; \label{eq:note1} \\
    \pi_f  P_{H_{f}^{-} \cap D^-}(f>0)  \mathbb{E}_{H_{f}^{-} \cap D^-}[f|f> 0] & = 2  \pi_f  P_{H_{f}^{-} \cap D^-}(f>0)  \lvert \mathbb{E}_{H_{f}^{-} \cap D^-}[f|f> 0] \rvert  \notag \\
    & - \pi_f  P_{H_{f}^{-} \cap D^-}(f>0)  \mathbb{E}_{H_{f}^{-} \cap D^-}[f|f>0]. \notag   \\
     & \leq 2  P_{H_{f}^{-} \cap D^-}(f>0)  \pi  c^{max}_{H_{f}^{-}}(f)  -\pi_f  P_{H_{f}^{-} \cap D^-}(f>0)  \mathbb{E}_{H_{f}^{-} \cap D^-}[f|f>0]. \label{eq:note2}
\end{align}

By assumption \ref{ass:distance} from the main text, fact \ref{eq:fact_max} and fact \ref{eq:fact_jensen} we have:
\begin{align*}
    & \mathbb{E}_{A \cap D^{+/-}}[c^*|f\leq 0 ,g\leq 0]  \leq \alpha  \mathbb{E}_{A \cap D^{+/-}}[ \max(-f, -g)^{\gamma} |f\leq 0 ,g\leq 0] \\
    & = \left( \frac{\alpha}{2^\gamma}\right )  (\mathbb{E}_{A \cap D^{+/-}}[(-f - g + |-f+g|)^{\gamma} |f\leq 0 ,g\leq 0]) 
     \leq  \left( \frac{\alpha}{2^\gamma}\right )  (\mathbb{E}_{A \cap D^{+/-}}[(-f - g + |-f+g|)|f\leq 0 ,g\leq 0])^{\gamma}.
\end{align*}

\begin{align*}
    & \mathbb{E}_{A \cap D^{+/-}}[c^*|f\leq 0 ,g > 0]  \leq \alpha  \mathbb{E}_{A \cap D^{+/-}}[ \max(-f, +g)^{\gamma} |f\leq 0 ,g > 0]  \\
    & = \left( \frac{\alpha}{2^\gamma}\right )  (\mathbb{E}_{A \cap D^{+/-}}[(-f + g + |-f-g|)^{\gamma} |f\leq 0 ,g > 0]) \leq  \left( \frac{\alpha}{2^\gamma}\right )  (\mathbb{E}_{A \cap D^{+/-}}[(-f + g + |-f-g|)|f\leq 0 ,g>0])^{\gamma}.
\end{align*}
Analogously for the remaining 2 terms.
Next, note that
\begin{align}
    & \mathbb{E}_{H_{f}^{-} \cup H_{g}^{-} \cap D^{+/-}}[f|f\leq 0 ,g\leq 0]  \leq \mathbb{E}_{H_{f}^{-} \cup H_{g}^{-} \cap D^{+/-}}[f|f\leq 0] = 
     \mathbb{E}_{H_{f}^{-} \cap D^{+/-}}[f|f\leq 0], \label{eq:simplification_1} \\
    &\mathbb{E}_{H_{f}^{-} \cup H_{g}^{-} \cap D^{+/-}}[g|f\leq 0 ,g\leq 0]  \leq \mathbb{E}_{H_{f}^{-} \cup H_{g}^{-} \cap D^{+/-}}[g|g\leq 0] =
     \mathbb{E}_{H_{g}^{-} \cap D^{+/-}}[g|g\leq 0] \label{eq:simplification_2},
\end{align}
where the first equality follows by assuming that $f$ and $g$ do not assign widely different predictions to the same input and the second equality follows since $H_{g}^{-}$ is not restricting $H_{f}^{-} \cup H_{g}^{-}$ and vice versa. Similarly, for the remaining terms.
Now, we go back to the expanded cost, use linearity of expectations, \eqref{eq:simplification_1}, \eqref{eq:simplification_2}, facts \ref{eq:fact_intersection_inequality} (second inequality) and \ref{eq:fact_probability_inequality} (first inequality) and upper bound it by:
\begin{align*}
    & \overline{cost}(f,g)_{H_{f}^{-} \cup H_{g}^{-}}  \leq  \pi^\gamma  \left( \frac{\alpha}{2^\gamma} \right)  \bigg[ \mathbb{E}_{A \cap D^{+}}[(-f - g + |-f+g||f\leq 0 ,g\leq 0]^{\gamma}  P_{A \cap D^{+}}(f\leq 0 ,g\leq 0)^\gamma \\
    & + \mathbb{E}_{A \cap D^{+}}[(-f + g + |-f-g||f\leq 0 ,g\leq 0]^{\gamma}  P_{A \cap D^{+}}(f\leq 0 ,g > 0)^{\gamma}  \\
    & + \mathbb{E}_{A \cap D^{+}}[(f + g + |f-g||f\leq 0 ,g\leq 0]^{\gamma}  P_{A \cap D^{+}}(f>0,g>0)^{\gamma}  \\
    & + \mathbb{E}_{A \cap D^{+}}[(f - g + |f+g||f\leq 0 ,g\leq 0]^{\gamma}  P_{A \cap D^{+}}(f>0,g\leq0)^{\gamma} \bigg] \\ 
    & +\pi_{-}^\gamma  \left( \frac{\alpha}{2^\gamma} \right) \bigg[ \mathbb{E}_{A \cap D^{-}}[(-f - g + |-f+g||f\leq 0 ,g\leq 0])^{\gamma}   P_{A \cap D^{-}}(f\leq 0 ,g\leq 0)^\gamma \\
    & + \mathbb{E}_{A \cap D^{-}}[(-f + g + |-f-g||f\leq 0 ,g\leq 0]^{\gamma}  P_{A \cap D^{-}}(f\leq 0 ,g > 0)^{\gamma}  \\
    & + \mathbb{E}_{A \cap D^{-}}[(f + g + |f-g||f\leq 0 ,g\leq 0]^{\gamma}  P_{A \cap D^{-}}(f>0,g>0)^{\gamma}  \\
    & + \mathbb{E}_{A \cap D^{-}}[(f - g + |f+g||f\leq 0 ,g\leq 0]^{\gamma}  P_{A \cap D^{-}}(f>0,g\leq0)^{\gamma} \bigg]
    \end{align*}
    \begin{align*}
    & \leq \pi^\gamma  \left( \frac{\alpha}{2^\gamma} \right)  \bigg[ \bigg(\mathbb{E}_{H_{f}^{-} \cap D^{+}}[-f|f\leq 0]  P_{H_{f}^- \cap D^{+}}(f\leq 0 ) + \mathbb{E}_{H_{g}^{-} \cap D^{+}}[-g|g\leq 0]  P_{H_g^- \cap D^{+}}(g\leq 0)] \\
    & + \mathbb{E}_{A \cap D^{+}}[|-f+g||f\leq 0 ,g\leq 0]  P_{A \cap D^{+}}(f\leq 0 ,g\leq 0) \bigg)^\gamma  \\ 
    & + \bigg(\mathbb{E}_{H_{f}^{-} \cap D^{+}}[-f|f\leq 0]  P_{H_{f}^- \cap D^{+}}(f\leq 0)  + \mathbb{E}_{H_{g}^{-} \cap D^{+}}[g|g>0]  P_{H_g^- \cap D^{+}}(g>0) \\
    & + \mathbb{E}_{A \cap D^{+}}[|-f-g||f\leq 0 ,g>0]  P_{A \cap D^{+}}(f\leq 0 ,g>0) \bigg)^\gamma \\ 
    & + \bigg(\mathbb{E}_{H_{f}^{-} \cap D^{+}}[f|f>0]  P_{H_{f}^- \cap D^{+}}(f>0) + \mathbb{E}_{H_{g}^{-} \cap D^{+}}[g|g>0]  P_{H_g^- \cap D^{+}}(g>0) \\
    & + \mathbb{E}_{H_{g}^- \cap D^{+}}[|f-g||f>0 ,g>0]  P_{A \cap D^{+}}(f>0 ,g> 0) \bigg)^\gamma \\
    & + \bigg(\mathbb{E}_{H_{f}^{-} \cap D^{+}}[f|f> 0]  P_{H_{f}^- \cap D^{+}}(f> 0) + \mathbb{E}_{H_{g}^{-} \cap D^{+}}[-g|g\leq 0]  P_{H_g^- \cap D^{+}}(g\leq 0) \\
    & + \mathbb{E}_{A \cap D^{+}}[|f+g||f>0 ,g\leq0]  P_{A \cap D^{+}}(f>0 ,g\leq0) \bigg)^\gamma \bigg] \\
    & + \pi_{-}^\gamma  \left( \frac{\alpha}{2^\gamma} \right)  \bigg[ \bigg(\mathbb{E}_{H_{f}^{-} \cap D^{+}}[-f|f\leq 0]  P_{H_{f}^- \cap D^{+}}(f\leq 0) + \mathbb{E}_{H_{g}^{-} \cap D^{+}}[-g|g\leq 0]  P_{H_g^- \cap D^{+}}(g\leq 0) \\
    & + \mathbb{E}_{A \cap D^{+}}[|-f+g||f\leq 0 ,g\leq 0]  P_{A \cap D^{+}}(f\leq 0 ,g\leq 0) \bigg)^\gamma \\ 
    & + \bigg(\mathbb{E}_{H_{f}^{-} \cap D^{-}}[-f|f\leq 0]  P_{H_{f}^- \cap D^{-}}(f\leq 0) + \mathbb{E}_{H_{g}^{-} \cap D^{-}}[g|g>0]  P_{H_g^- \cap D^{+}}(g>0) \\
    & + \mathbb{E}_{A \cap D^{-}}[|-f-g||f\leq 0 ,g>0]  P_{A \cap D^{+}}(f\leq 0 ,g>0) \bigg)^\gamma \\ 
    & + \bigg(\mathbb{E}_{H_{f}^{-} \cap D^{-}}[f|f>0]  P_{H_{f}^- \cap D^{-}}(f>0) + \mathbb{E}_{H_{g}^{-} \cap D^{-}}[g|g>0]  P_{H_g^- \cap D^{-}}(g>0) \\
    & + \mathbb{E}_{H_{g}^- \cap D^{-}}[|f-g||f>0 ,g>0]  P_{A \cap D^{-}}(f>0 ,g> 0) \bigg)^\gamma  \\ 
    & + \bigg(\mathbb{E}_{H_{f}^{-} \cap D^{-}}[f|f> 0]  P_{H_{f}^- \cap D^{-}}(f> 0) + \mathbb{E}_{H_{g}^{-} \cap D^{-}}[-g|g\leq 0]  P_{H_g^- \cap D^{-}}(g\leq 0) \\
    & + \mathbb{E}_{A \cap D^{-}}[|f+g||f>0 ,g\leq0]  P_{A \cap D^{-}}(f>0 ,g\leq0) \bigg)^\gamma \bigg] 
\end{align*}

Next we apply lemma \ref{eq:lemma_fawzi} with $n=8$, use \eqref{eq:note1} and \eqref{eq:note2} for the last equality and obtain:
\begin{align*}
& \leq 8^{1-\gamma} \left( \frac{\alpha}{2^\gamma} \right) \bigg[ \pi \bigg(\bigg(\mathbb{E}_{H_{f}^{-} \cap D^{+}}[-f|f\leq 0]  P_{H_{f}^- \cap D^{+}}(f\leq 0 ) + \mathbb{E}_{H_{g}^{-} \cap D^{+}}[-g|g\leq 0]  P_{H_g^- \cap D^{+}}(g\leq 0)] \\
    & + \mathbb{E}_{A \cap D^{+}}[|-f+g||f\leq 0 ,g\leq 0]  P_{A \cap D^{+}}(f\leq 0 ,g\leq 0) \bigg)  \\ 
    & + \bigg(\mathbb{E}_{H_{f}^{-} \cap D^{+}}[-f|f\leq 0]  P_{H_{f}^- \cap D^{+}}(f\leq 0) + \mathbb{E}_{H_{g}^{-} \cap D^{+}}[g|g>0]  P_{H_g^- \cap D^{+}}(g>0) \\ & + \mathbb{E}_{A \cap D^{+}}[|-f-g||f\leq 0 ,g>0]  P_{A \cap D^{+}}(f\leq 0 ,g>0) \bigg) \\
    & + \bigg(\mathbb{E}_{H_{f}^{-} \cap D^{+}}[f|f>0]  P_{H_{f}^- \cap D^{+}}(f>0)  + \mathbb{E}_{H_{g}^{-} \cap D^{+}}[g|g>0]  P_{H_g^- \cap D^{+}}(g>0) \\
    & + \mathbb{E}_{H_{g}^- \cap D^{+}}[|f-g||f>0 ,g>0]  P_{A \cap D^{+}}(f>0 ,g> 0) \bigg)  \\ 
    & + \bigg(\mathbb{E}_{H_{f}^{-} \cap D^{+}}[f|f> 0]  P_{H_{f}^- \cap D^{+}}(f> 0)  + \mathbb{E}_{H_{g}^{-} \cap D^{+}}[-g|g\leq 0]  P_{H_g^- \cap D^{+}}(g\leq 0) \\
    & + \mathbb{E}_{A \cap D^{+}}[|f+g||f>0 ,g\leq0]  P_{A \cap D^{+}}(f>0 ,g\leq0) \bigg)\bigg) \\ 
    & + \pi_{-}  \bigg( \bigg(\mathbb{E}_{H_{f}^{-} \cap D^{+}}[-f|f\leq 0]  P_{H_{f}^- \cap D^{+}}(f\leq 0 )  + \mathbb{E}_{H_{g}^{-} \cap D^{+}}[-g|g\leq 0]  P_{H_g^- \cap D^{+}}(g\leq 0)] \\
    & + \mathbb{E}_{A \cap D^{+}}[|-f+g||f\leq 0 ,g\leq 0]  P_{A \cap D^{+}}(f\leq 0 ,g\leq 0) \bigg)  \\ 
    & + \bigg(\mathbb{E}_{H_{f}^{-} \cap D^{-}}[-f|f\leq 0]  P_{H_{f}^- \cap D^{-}}(f\leq 0) + \mathbb{E}_{H_{g}^{-} \cap D^{-}}[g|g>0]  P_{H_g^- \cap D^{+}}(g>0) \\
    & + \mathbb{E}_{A \cap D^{-}}[|-f-g||f\leq 0 ,g>0]  P_{A \cap D^{+}}(f\leq 0 ,g>0) \bigg) \\
    & + \bigg(\mathbb{E}_{H_{f}^{-} \cap D^{-}}[f|f>0]  P_{H_{f}^- \cap D^{-}}(f>0) +  \mathbb{E}_{H_{g}^{-} \cap D^{-}}[g|g>0]  P_{H_g^- \cap D^{-}}(g>0) \\
    & + \mathbb{E}_{H_{g}^- \cap D^{-}}[|f-g||f>0 ,g>0]  P_{A \cap D^{-}}(f>0 ,g> 0) \bigg)  \\ 
    & + \bigg(\mathbb{E}_{H_{f}^{-} \cap D^{-}}[f|f> 0]  P_{H_{f}^- \cap D^{-}}(f> 0)  + \mathbb{E}_{H_{g}^{-} \cap D^{-}}[-g|g\leq 0]  P_{H_g^- \cap D^{-}}(g\leq 0) \\
    & + \mathbb{E}_{A \cap D^{-}}[|f+g||f>0 ,g\leq0]  P_{A \cap D^{-}}(f>0 ,g\leq0) \bigg)\bigg]^\gamma \end{align*} 
     \begin{align*}
    &=  8^{1-\gamma} \left( \frac{\alpha}{2^\gamma} \right) \bigg[ \pi  \bigg( 2  \mathbb{E}_{H_{f}^{-} \cap D^{+}}[-f|f\leq 0]  P_{H_{f}^- \cap D^{+}}(f\leq 0 ) + 2  \mathbb{E}_{H_{g}^{-} \cap D^{+}}[-g|g\leq 0]  P_{H_g^- \cap D^{+}}(g\leq 0)] \\
     & + 2   \mathbb{E}_{H_{f}^{-} \cap D^{+}}[f|f>0]  P_{H_{f}^- \cap D^{+}}(f>0)  + 2   \mathbb{E}_{H_{g}^{-} \cap D^{+}}[g|g>0]  P_{H_g^- \cap D^{+}}(g>0) \bigg) \\ 
    & + \pi_{-}  \bigg( 2  \mathbb{E}_{H_{f}^{-} \cap D^{-}}[-f|f\leq 0]  P_{H_{f}^- \cap D^{-}}(f\leq 0 )  + 2  \mathbb{E}_{H_{g}^{-} \cap D^{-}}[-g|g\leq 0]  P_{H_g^- \cap D^{-}}(g\leq 0)] \\ 
    & + 2   \mathbb{E}_{H_{f}^{-} \cap D^{-}}[f|f>0]  P_{H_{f}^- \cap D^{-}}(f>0)  + 2   \mathbb{E}_{H_{g}^{-} \cap D^{-}}[g|g>0]  P_{H_g^- \cap D^{-}}(g>0)\bigg) + \mathbb{E}_{A}[|f-g|]  \bigg]^\gamma \\
    & = \alpha  8^{1-\gamma}  \bigg[2  \bigg(\frac{2  [ R_{H_f^-}(f)  c^{max}_{H_f^-}(f) +  R_{H_g^-}(g)  c^{max}_{H_g^-}(g)]}{2}  \\
    & + \frac{\pi_f   \mathbb{E}_{H_f^- \cap D^+}[f] + \pi_g   \mathbb{E}_{H_g^- \cap D^+}[g]}{2}  - \frac{(1-\pi_f)   \mathbb{E}_{H_f^- \cap D^-}[f] + (1-\pi_g)   \mathbb{E}_{H_g^- \cap D^-}[g]}{2}\bigg) + \mathbb{E}_{H_f^- \cup H_g^-}[|f-g|]\bigg]^\gamma,
\end{align*}
where
\begin{align*}
\mathbb{E}_{A}[|f-g|] &:= \pi  \bigg(  \mathbb{E}_{A \cap D^{+}}[|-f+g||f\leq 0 ,g\leq 0]   P_{A \cap D^{+}}(f\leq 0 ,g\leq 0) )\\
& +  \mathbb{E}_{A \cap D^{+}}[|-f-g||f\leq 0 ,g>0]  P_{A \cap D^{+}}(f\leq 0 ,g>0) ) \\
& +  \mathbb{E}_{A \cap D^{+}}[|f-g||f>0 ,g>0]  P_{A \cap D^{+}}(f>0 ,g> 0) )  \\
& + \mathbb{E}_{A \cap D^{+}}[|f+g||f>0 ,g\leq0]  P_{A \cap D^{+}}(f>0 ,g\leq0) ) \bigg) \\
& + \pi_{-}  \bigg( \mathbb{E}_{A \cap D^{-}}[|-f+g||f\leq 0 ,g\leq 0] P_{A \cap D^{-}}(f\leq 0 ,g\leq 0) ) \\
& + \mathbb{E}_{A \cap D^{-}}[|-f-g||f\leq 0 ,g>0]  P_{A \cap D^{-}}(f\leq 0 ,g>0)) \\
& +  \mathbb{E}_{A \cap D^{-}}[|f-g||f>0 ,g>0]  P_{A \cap D^{-}}(f>0 ,g> 0)) \\
& + \mathbb{E}_{A \cap D^{-}}[|f+g||f>0 ,g\leq0]  P_{A \cap D^{-}}(f>0 ,g\leq0))\bigg).
\end{align*}
and (using \eqref{eq:note1} and \eqref{eq:note2}) to rewrite the first line and fact \ref{eq:fact_omission_rates} to establish the inequality. 
\begin{align*}
& 2  \pi  \bigg[-\mathbb{E}_{H_f^-\cap D^+}[f|f\leq0]  P_{H_f^-\cap D^+}(f|f\leq0)  + \mathbb{E}_{H_f^-\cap D^+}[f|f>0]  P_{H_f^-\cap D^+}(f|f>0)\bigg] \\
& + 2  \pi_{-}   \bigg[-\mathbb{E}_{H_f^-\cap D^-}(f|f\leq0)  P_{H_f^-\cap D^-}(f|f\leq0) + \mathbb{E}_{H_f^-\cap D^-}[f|f>0]  P_{H_f^-\cap D^-}(f|f>0)\bigg] \\
& \leq 2  \bigg[2 \pi_f   P_{H_f^-\cap D^+}(f|f\leq0)  c_{H^-_f}^{max}(f)  - \pi_f   P_{H_f^-\cap D^+}(f\leq0)  \mathbb{E}_{H_f^-\cap D^+}[f|f>0] +\pi_f   \mathbb{E}_{H_f^-\cap D^+}(f|f>0)  P_{H_f^-\cap D^+}(f>0)  \\
& - \bigg( - (1-\pi_f)   2  P_{H_f^-\cap D^-}(f|f>0)  c_{H^-_f}^{max}(f)  + (1-\pi_f)   P_{H_f^-\cap D^-}(f|f>0)  \mathbb{E}_{H_f^-\cap D^-}[f|f>0] \\
& + (1-\pi_f)   P_{H_f^-\cap D^-}(f|f\leq0)  \mathbb{E}_{H_f^-\cap D^-}[f|f\leq0] \bigg) \bigg]  \\
& = 2 \bigg[2   R_{H_f^-}(f)  c_{H^-_f}^{max}(f)  + \pi_f   \mathbb{E}_{H_f^-\cap D^+}[f] - (1-\pi_{f})  \mathbb{E}_{H_f^-\cap D^-}[f]\bigg].
\end{align*}
and 
\begin{align*}
     R_{H_f^-}(f) &:=  \bigg[\pi_f  P_{H_f^-\cap D^+}(f|f\leq0)  + (1-\pi_f)   P_{H_f^-\cap D^-}(f|f>0) \bigg].  \\
\end{align*}
\end{proof}

\section{Proof of proposition \ref{proposition:relation_negative_surprise}} \label{appendix:relation_negative_surprise}

In essence, we wish to identify the conditions under which $\overline{s}(f,g)_{S} \lesseqqgtr \overline{s}(f,g)_{D}$.

We first note that the following result is immediate from proposition \ref{proposition:oracle_cost}.
\begin{corollary}
\begin{equation*}
\mathbb{E}_{H_f^-}[c^*(f)]_{S} \leq  \mathbb{E}_{H_f^-}[c^*(f)]_{D}.
\end{equation*}
\label{corollary:expectations}
\end{corollary}

Using lemma \ref{lemma:ustun}, we can lower bound (3) from the main text as follows: \begin{align*}
&\bigg( \mathbb{E}_{H_{f}^{-}}[c^*(f)]_M
+\mathbb{E}_{H_{g}^{-}}[c^*(g)]_M \\
& + \alpha \mathbb{E}_{H_f^- \cup H_g^-}[|f(x)-g(x)|]_M \bigg)^\gamma \bigg(8 \cdot \alpha \bigg)^{1-\gamma} \\
& \leq \mathbb{E}_{H_{f}^{-} \cup H_{g}^{-}}[c^*(f,g)]_M.
\end{align*}

For simplicity of the statement we assume that $\gamma=1$. We can now find an upper bound for the expected inverse cost of negative surprise under method $M = \{S, D\}$:
\begin{equation*}
    \overline{s}(f,g)_M = \frac{\mathbb{E}_{H_{f}^{-}}[c^*(f)]_M}{\mathbb{E}_{H_{f}^{-} \cup H_{g}^{-}}[c^*(f,g)]_M} \leq \frac{\mathbb{E}_{H_{f}^{-}}[c^*(f)]_M}{\mathbb{E}_{H_{f}^{-}}[c^*(f)]_M
+\mathbb{E}_{H_{g}^{-}}[c^*(g)]_M + \alpha \mathbb{E}_{H_f^- \cup H_g^-}[|f(x)-g(x)|]_M}.
\end{equation*}
For simplicity, by corollary \ref{corollary:expectations} we can set:
\begin{align*}
    \mathbb{E}_{H_{f}^{-}}[c^*(f)]_{S} + \delta_f & = \mathbb{E}_{H_{f}^{-}}[c^*(f)]_{D}, \\
    \mathbb{E}_{H_{g}^{-}}[c^*(g)]_{S} + \delta_g & = \mathbb{E}_{H_{g}^{-}}[c^*(g)]_{D} ,
\end{align*}
for some $\delta_f, \delta_g > 0$.

\textbf{Proposition}. Suppose $\overline{s}(f,g)_{S} > \overline{s}(f,g)_{D}$ and $\mathbb{E}_{H_f^- \cup H_g^-}[|f(x)-g(x)|]_{S} = \mathbb{E}_{H_f^- \cup H_g^-}[|f(x)-g(x)|]_{D} := k$ hold, then we must have:
\begin{equation*}
    \frac{\mathbb{E}_{H_{g}^{-}}[c^*(g)]_{D}}{\mathbb{E}_{H_{f}^{-}}[c^*(f)]_{D}} < \frac{\mathbb{E}_{H_{g}^{-}}[c^*(g)]_{S}}{\mathbb{E}_{H_{f}^{-}}[c^*(f)]_{S}}.
\end{equation*}

Using the definition $\overline{s}(f,g)_{M}$ and the stated assumptions we can write $\overline{s}(f,g)_{S} > \overline{s}(f,g)_{D}$ as follows:
\begin{align*}
    1+ \underbrace{\frac{\mathbb{E}_{H_{g}^{-}}[c^*(g)]_{S} + \delta_g}{\mathbb{E}_{H_{f}^{-}}[c^*(f)]_{S} + \delta_f}}_{b_1} + \underbrace{\frac{\alpha k}{\mathbb{E}_{H_{f}^{-}}[c^*(f)]_{S} +\delta_f}}_{a_1} < 1 + \underbrace{\frac{\mathbb{E}_{H_{g}^{-}}[c^*(g)]_{S} }{\mathbb{E}_{H_{f}^{-}}[c^*(f)]_{S}}}_{b_2} + \underbrace{\frac{\alpha k}{\mathbb{E}_{H_{f}^{-}}[c^*(f)]_{S}}}_{a_2}
\end{align*}
Note that $a_1<a_2$ for $\delta_f > 0$. So the terms that remain to be checked are $b_1$ and $b_2$. Hence, we obtain
\begin{equation*}
    \frac{\mathbb{E}_{H_{g}^{-}}[c^*(g)]_{D}}{\mathbb{E}_{H_{f}^{-}}[c^*(f)]_{D}} < \frac{\mathbb{E}_{H_{g}^{-}}[c^*(g)]_{S}}{\mathbb{E}_{H_{f}^{-}}[c^*(f)]_{S}},
\end{equation*}
as desired.

\section{Other Prerequisites}

\begin{fact} \label{eq:fact_max}
$max(a,b) = \frac{1}{2}(a+b+|a-b|)$.
\end{fact}

\begin{fact} \label{eq:fact_intersection_inequality}
$P(A \cap B) \leq P(A);  P(A \cap B) \leq P(B)$.
\end{fact}

\begin{fact} \label{eq:fact_jensen}
$E[X^{\gamma}] \leq E[X]^{\gamma}$ for $0\leq \gamma \leq1$ (by Jensen's inequality).
\end{fact}

\begin{fact} \label{eq:fact_probability_inequality}
$P(X) \leq P(X)^{\gamma}$ \text{for} $0\leq \gamma \leq1$.
\end{fact}

\begin{fact} \label{eq:fact_omission_rates}
Note that $\pi \leq \pi_f$ and $\pi \leq \pi_g$.
\end{fact}

We state the following lemmata without proof.
\begin{lemma}[\citet{ustun2019actionable}]
For $\gamma = 1$, the expected cost of counterfactual explanations under model $f$, $\mathbb{E}_{H_f^-}[c^*(f,x)]$, is bounded from above such that:
\begin{equation*}
\overline{cost}_{H_f^-}(f) \leq \alpha \bigg(\pi_f \cdot c_{D^+}(f) - (1-\pi_f) \cdot c_{D^-}(f) + 2\cdot c_{H_f^-}^{max} \cdot R_{H_f^-}(f)  \bigg).
\end{equation*}
\label{lemma:ustun}
\end{lemma}

\begin{lemma}[\cite{fawzi2018analysis}]
Let $z_1, ..., z_n$ be non-negative real numbers, and let $0 \leq \gamma \leq 1$. Then
\begin{equation*} \label{eq:lemma_fawzi}
    \sum_{i=1}^n z_i^{\gamma} \leq n^{1-\gamma} \cdot \left(\sum_{i=1}^n z_i \right)^\gamma.
\end{equation*}
\end{lemma}
\end{appendices}

\end{document}